\documentclass[10pt,twocolumn,letterpaper]{article}

\usepackage{cvpr}              

\usepackage{amsmath}
\usepackage{amssymb}
\usepackage{algorithm}
\usepackage[noend]{algpseudocode}

\definecolor{cvprblue}{rgb}{0.21,0.49,0.74}
\usepackage[pagebackref,breaklinks,colorlinks,allcolors=cvprblue]{hyperref}
\usepackage{booktabs}
\usepackage{multirow}
\usepackage[table]{xcolor}
\usepackage{float}
\usepackage{placeins}
\usepackage{comment}
\usepackage[accsupp]{axessibility}

\definecolor{colorfirst}{HTML}{ffb2b3}
\definecolor{colorsecond}{HTML}{ffd9b6}
\definecolor{colorthird}{HTML}{ffffb9}

\def\paperID{4978} 
\def\confName{CVPR}
\def\confYear{2026}

\title{ExMesh: Explicit Mesh Reconstruction with Topology Adaptation}

\vspace{-2mm}

\author{
Chuanjin Fan$^{1}$\; Lifan Wu$^{1}$\; Wenjie Chang$^{1}$\; Hanzhi Chang$^{1}$\; Wenfei Yang$^{1}$\dag\; Tianzhu Zhang$^{1,2}$ \\
$^{1}$University of Science and Technology of China \\
$^{2}$National Key Laboratory of Deep Space Exploration, Deep Space Exploration Laboratory
}

\begin{document}
\twocolumn[{
\renewcommand\twocolumn[1][]{#1}
\maketitle

\vspace{-8mm}
\begin{center}
    \captionsetup{type=figure}
    \includegraphics[width=0.98\linewidth]{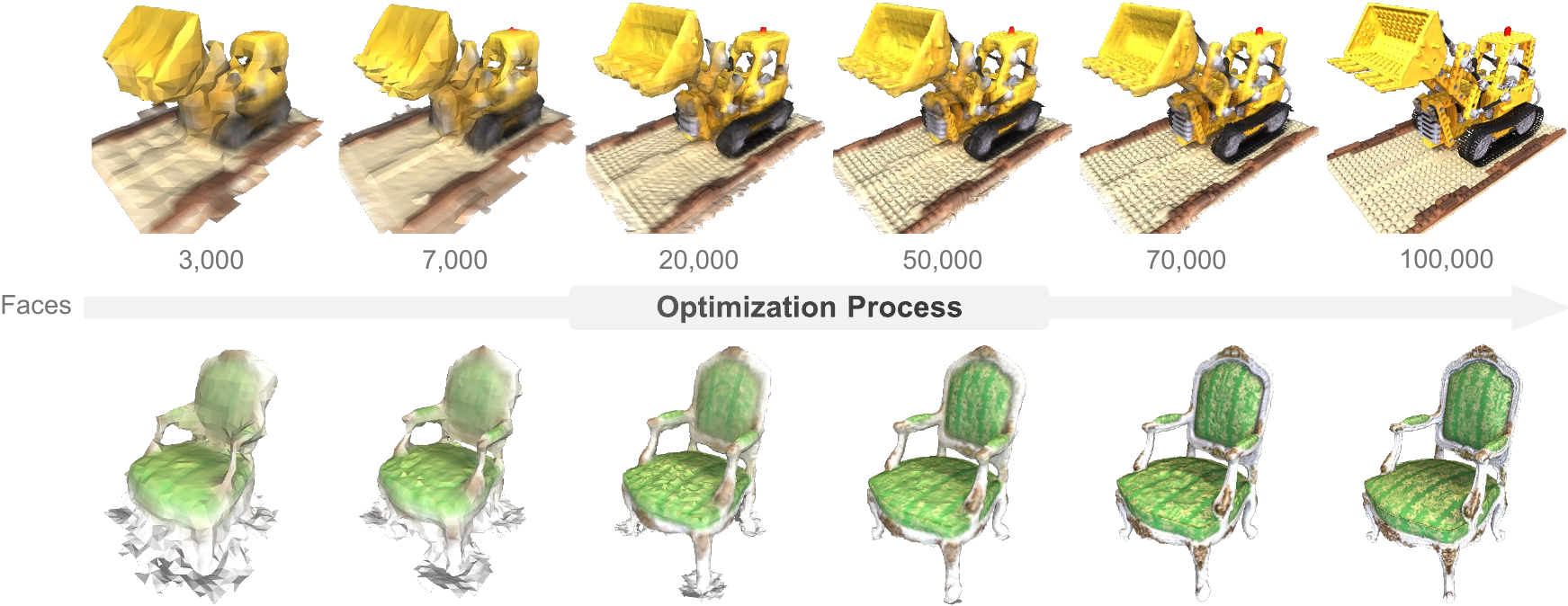}
    \vspace{-1mm}
    \captionof{figure}{The coarse-to-fine optimization process of our ExMesh framework. Driven by adaptive vertex splitting and merging, together with decoupled geometry-texture optimization, our method refines a coarse mesh into a high-fidelity result. This approach achieves a superior trade-off between reconstruction quality, computational efficiency, and a concise mesh structure.}
    \label{fig:teaser}
\end{center}
\vspace{2mm}
}]

\begin{abstract}
Reconstructing surface meshes from multi-view images has remained a core challenge in recent years. Most existing methods, whether implicit or explicit, depend on intermediate representations and post-processing steps like Marching Cubes or TSDF fusion, often resulting in artifacts and fragmented geometry. Directly optimizing explicit meshes is a promising approach. However, it presents two critical challenges. The first is how to adaptively refine mesh topology to capture detail without introducing degenerate faces. The second is how to maintain consistent UV coordinates for high-fidelity texturing as the mesh structure evolves. To overcome these, we propose ExMesh, a novel framework that directly optimizes explicit meshes by integrating differentiable optimization with discrete topology updates. Specifically, we introduce an adaptive vertex splitting and merging strategy, along with real-time UV maintenance, to enable coarse-to-fine optimization while preserving geometric integrity. To our knowledge, ExMesh is the first framework to seamlessly integrate discrete topology operations into a continuous differentiable optimization pipeline. Extensive experiments demonstrate that ExMesh achieves a balance among accuracy, computational efficiency, and mesh conciseness.
\end{abstract}
\vspace{-4mm}    
\section{Introduction}
\label{sec:intro}
Surface meshes, as a core form of explicit geometric representation, hold irreplaceable value across various domains, including virtual/augmented reality, digital twins, and robotics. Their representation not only supports real-time rasterized rendering but can also be directly used for scene editing and interaction~\cite{hughes1990cgpp,akenine2018rtr,catmull1978bspline,sorkine2007asrigid,terzopoulos1987elastically,baraff1998largesteps}. However, reconstructing high-fidelity meshes from multi-view images faces a core challenge, which lies in balancing three aspects: accuracy, computational efficiency, and mesh conciseness.

Recently, with the development of Neural Radiance Fields, implicit methods~\cite{yariv2021volsdf,wang2021neus,tancik2023neuralangelo,fu2022geoneus} represent scenes as continuous fields learned by MLPs. When extracting meshes, however, these methods must discretize the field into high-resolution voxels and apply Marching Cubes~\cite{lorensen1987marchingcubes}. This process often results in blurred sharp edges, loss of thin structures, and extremely long training time. Explicit methods using discrete Gaussian primitives (or their variants)~\cite{kerbl2023gaussian,huang2024twodgs,held2025trianglesplatting,yu2024gof,chen2024pgsr,zhang2025qgs} have improved training and rendering efficiency. Nevertheless, they still require post-processing, such as TSDF fusion~\cite{curless1996volumetric}, to convert the unstructured Gaussians into meshes, often producing a large number of discrete fragments. Meanwhile, mesh-driven methods based on hybrid representations have emerged~\cite{munkberg2022nvdiffrec,shen2021dmtet,shen2023flexicubes,wang2018pixel2mesh,gao2019tmnet,yang2025imlssplatting,li2025geosvr}. Instead of directly optimizing meshes, these methods use SDFs, point clouds, or voxels as the primary optimization targets, which increases framework complexity and can compromise geometric accuracy. Moreover, if texture is explicitly stored on the geometric carrier, its density must be increased to preserve detail, leading to a significant rise in face count~\cite{wang2021neus,tancik2023neuralangelo,su2023sugar,yang2025imlssplatting}. While coordinate-based MLPs~\cite{munkberg2022nvdiffrec} can decouple texture resolution from geometry, they still require complex post-training baking. These drawbacks make it difficult to maintain both high accuracy and simplicity, highlighting the necessity of moving beyond intermediate representations.

Directly optimizing position and texture is the ideal path to bypass aforementioned drawbacks~\cite{held2025meshsplatting}, but it also faces severe technical challenges. At the geometry processing level, this path faces the dual challenge of how to adaptively refine the mesh and how to maintain its structural integrity. Intuitively, an effective strategy would allocate a higher density of faces to geometrically complex regions and lower density to flatter areas. However, existing global uniform refinement strategies directly perform quad-splitting on all faces~\cite{munkberg2022nvdiffrec,NeuralPull,wang2018pixel2mesh,dou2024micromesh}, resulting in significant face redundancy in flat regions and insufficient detail recovery in complex areas. On the other hand, vertex displacement during optimization can easily produce degenerate faces, leading to substantial degradation of geometric accuracy. At the texture representation level, a fundamental constraint is that the quality of texture detail is highly dependent on face density. If texture is directly stored on vertices or faces, even geometrically simple surfaces with complex textures require a massive number of faces to represent texture detail. Although using a decoupled UV map is the ideal solution, real-time maintenance of UV coordinates under changing topology remains an unsolved challenge~\cite{munkberg2022nvdiffrec,yang2022neumesh}.

In this paper, we propose a novel mesh reconstruction framework that integrates differentiable optimization with discrete topology updates. By updating mesh structure during optimization, our method directly produces a renderable and editable mesh without intermediate representations or post-processing. Specifically, to tackle the challenges at the geometry processing level, we design an adaptive vertex splitting and merging strategy. This strategy splits vertices in complex geometric regions with high curvature and large gradients to accurately recover details. Meanwhile, it merges vertices on redundant and degenerate faces generated during optimization, which effectively controls the face count and ensures geometric integrity. Furthermore, to break constraints at the texture representation level, we propose a method for real-time UV maintenance by updating UV coordinates alongside each topology change. By jointly optimizing vertex positions with a UV map, our method preserves texture details while significantly reducing the dependence on face count. To our knowledge, ExMesh is the first approach to achieve direct mesh optimization with real-time adaptive refinement, providing a new technical paradigm for mesh reconstruction.

In summary, our contributions are as follows:
\begin{itemize}
  \item We propose ExMesh, a novel end-to-end mesh reconstruction framework that integrates explicit optimization with discrete topology updates, directly producing structurally complete and editable meshes.
  \item An adaptive vertex splitting and merging strategy that achieves coarse-to-fine mesh refinement while effectively removing redundant and degenerate faces.
  \item A mechanism for real-time UV maintenance, enabling decoupled geometry and texture during topology updates for high-fidelity texturing on concise meshes.
\end{itemize}

\section{Related Work}
\label{sec:related_work}

\begin{figure*}[t]
    \centering
    \includegraphics[width=0.95\linewidth]{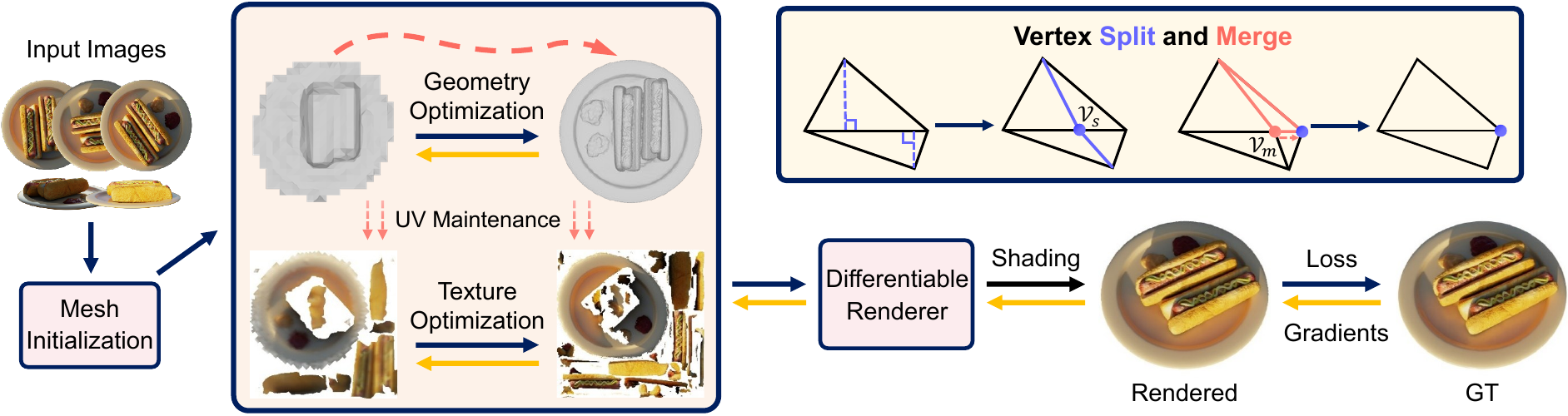}
    \vspace{-2mm}
    \caption{Overview of our ExMesh framework. We utilize a differentiable renderer to compute a photometric loss between the rendered image and the ground-truth (GT) image. The gradients are back-propagated to drive two decoupled optimization loops: (1) Geometry Optimization for vertex positions and (2) Texture Optimization for a separate UV map. The continuous geometry optimization is interleaved with discrete vertex split and merge operations to adaptively refine the mesh topology from coarse to fine.}
    \label{fig:method}
    \vspace{-2mm}
\end{figure*}

\subsection{Implicit Neural Surface Reconstruction}
Implicit neural representations, particularly Neural Radiance Fields (NeRF) \cite{mildenhall2020nerf}, have achieved impressive results in novel view synthesis and photorealistic rendering. However, the volume density field learned by NeRF makes it difficult to extract smooth and fine-grained surfaces. To address this, subsequent works use Signed Distance Functions \cite{yariv2020multiview,shang2020meshsdf,niemeyer2020dvr} to implicitly represent geometry. Among these, NeuS \cite{wang2021neus}, UNISURF \cite{oechsle2021unisurf} and VolSDF \cite{yariv2021volsdf} proposed SDF-to-density transformations, enabling optimization of SDF field via a photometric loss. Recent methods further improve quality and efficiency by introducing multi-resolution hash grids~\cite{mueller2022instantngp}, numerical gradients~\cite{tancik2023neuralangelo}, and geometric priors such as SfM point clouds~\cite{fu2022geoneus} to constrain SDF optimization. Despite advances in reconstruction quality, these methods remain indirect, relying on intermediate representations and post-processing (e.g., Marching Cubes~\cite{lorensen1987marchingcubes}) to extract meshes, which is time-consuming and often blurs sharp edges or loses thin structures.

\subsection{Gaussian Splatting-based Reconstruction}
To overcome the high training and rendering costs of implicit methods, a series of explicit methods have been proposed. Among them, 3D Gaussian Splatting \cite{kerbl2023gaussian} is a pioneering work that uses millions of anisotropic 3DGS primitives to represent the scene. Paired with a differentiable rasterizer, it achieves fast training and real-time rendering, with rendering quality surpassing NeRF in many aspects. Subsequent works explored other explicit primitives, such as 2DGS \cite{huang2024twodgs}, which flattens 3DGS into disks, and Triangle Splatting \cite{held2025trianglesplatting}, which directly optimizes a ``triangle soup''.

Although 3DGS and its variants have achieved great success in rendering, their primitives are essentially an unstructured point cloud, not directly editable manifold meshes. Therefore, to extract surface from these primitives, subsequent works \cite{su2023sugar,chen2024pgsr,zhang2025qgs,yu2024gof,huang2024twodgs,lyu2024_3dgsr,dai2024gaussiansurfels,wolf2024gs2mesh,chang2025meshsplat,stuart2025gs2pc} also rely on post-processing steps. These methods typically produce meshes using TSDF fusion \cite{curless1996volumetric,newcombe2011kinectfusion} or Poisson reconstruction \cite{kazhdan2013screened}. Although more detailed than Marching Cubes \cite{lorensen1987marchingcubes}, it still results in a large number of discrete mesh fragments and noise artifacts. Whether using implicit fields or Gaussian primitives, this indirect pipeline remains a core obstacle to high-quality and concise mesh reconstruction.

\subsection{Mesh-driven Reconstruction}
To circumvent the indirect pipeline, a series of works began exploring mesh-driven reconstruction. This was enabled by the development of differentiable rasterizers \cite{laine2020modular,liu2019softras,kato2018neural,chen2019idr}, which allow gradients to flow directly from image loss to mesh attributes. However, there remain two core challenges, the first of which is topology control. Early methods \cite{kato2018neural,liu2019softras} typically require a fixed topology template \cite{wang2018pixel2mesh,gao2019tmnet}, limiting the ability of geometry expressiveness. While some works explore topology changes via continuous remeshing~\cite{palfinger2022continuous} or probabilistic predictions~\cite{son2024dmesh,son2025dmeshpp}, they often lack adaptive, gradient-driven detail recovery. To support structural changes, later methods typically optimize intermediate carriers. Among them, Nvdiffrec~\cite{munkberg2022nvdiffrec} and FlexiCubes~\cite{shen2023flexicubes} optimize continuous SDF fields, while IMLS-Splatting~\cite{yang2025imlssplatting} and GeoSVR~\cite{li2025geosvr} use point clouds or voxels. Although these methods allow topology to change during training, their optimization target remains intermediate carriers rather than explicit meshes, resulting in constrained flexibility, increased complexity, and precision bias.

The second challenge is texture representation. Directly coupling texture with the geometric carrier~\cite{yang2025imlssplatting,li2025geosvr} requires a much denser carrier to capture high-frequency details, causing face count inflation. The ideal path is to use a decoupled texture map. Nvdiffrec \cite{munkberg2022nvdiffrec} addresses this by introducing a coordinate-based MLP texture, enabling high-frequency details on low-poly meshes. However, this implicit texture lacks direct editability and requires a post-training baking step to extract a standard UV map. Furthermore, its strategy requires freezing topology mid-training to preserve mapping continuity. Consequently, existing mesh-driven methods are caught in a dilemma: either achieve texture decoupling but sacrifice adaptive topology; or achieve dynamic topology at the expense of texture decoupling.
\section{Method}
\label{sec:method}

\subsection{Overview}
We propose ExMesh, a framework for reconstructing high-fidelity explicit meshes from multi-view images $\{I\}$. As shown in Figure~\ref{fig:method}, ExMesh takes an initial coarse mesh $\mathcal{M}_{init}$ as input, then iteratively optimizes its geometry and texture directly. In each iteration, we use a differentiable renderer \cite{laine2020modular} to render an image $\hat{I}$ for the current view, which is then compared with the ground-truth image $I$. Gradients from the loss are back-propagated to jointly optimize two components that are decoupled by design: (1) Geometry Optimization, which updates the 3D positions $\mathbf{V}$ of the vertices; and (2) Texture Optimization, which refines a separate UV map $\mathbf{T}$. To enable coarse-to-fine reconstruction, we interleave continuous gradient optimization with periodic vertex splitting and merging (Sec.~\ref{sec:split}, Sec.~\ref{sec:merge}), enabling dynamic mesh refinement and redundancy removal.

\begin{figure}[t]
    \centering
    \includegraphics[width=0.85\linewidth]{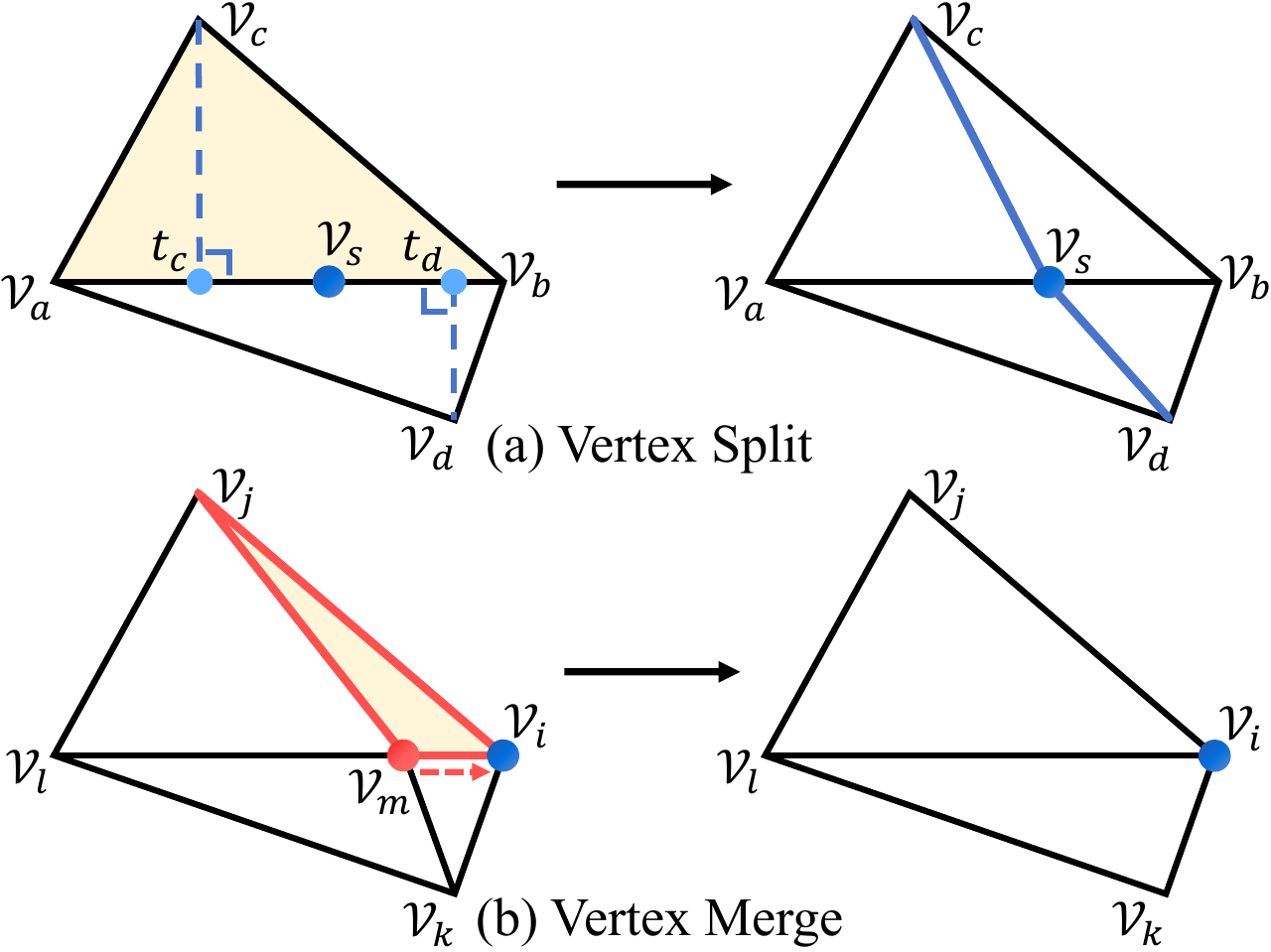}
    \vspace{-2mm}
    \caption{Our discrete topological operations. (a) Vertex Split: A new vertex $v_s$ is inserted to split the edge $(v_a, v_b)$, replacing 2 faces with 4 new ones. (b) Vertex Merge: Vertex $v_m$ is merged into $v_i$ by collapsing the edge $(v_i, v_m)$, simplifying the local topology.}
    \label{fig:vertex_split_merge}
    \vspace{-4mm}
\end{figure}

\subsection{Adaptive Vertex Splitting}
\label{sec:split}
Our topology update strategy borrows the idea of vertex splitting from mesh generation \cite{zhang2025vertexregen}, and combines it with a density control strategy from 3DGS \cite{kerbl2023gaussian}, allowing geometric complexity to be added adaptively. Vertex splitting process consists of two stages: criterion and operation.

\noindent\textbf{Splitting Criterion.} Whether a face is worth refining depends on two perspectives: the significance of its optimization error and the complexity of its geometry. In terms of optimization dynamics, the first metric we track is the vertex EMA gradient magnitude ($\mathcal{G}_v$), which uses an Exponential Moving Average (EMA) to accumulate the position gradient norm received by each vertex $\mathbf{V}$ during optimization. A high $\mathcal{G}_v$ indicates the vertex is located in a region requiring significant geometric adjustment. Its update formula is:

\vspace{-1mm}
\begin{equation}
\mathcal{G}_{v}^{(t)} = (1-\beta_g) \mathcal{G}_{v}^{(t-1)} + \beta_g \lVert \nabla_{\mathbf{V}}^{(t)} \rVert
\end{equation}

\noindent where $t$ denotes the current iteration step, $\beta_g$ is the EMA decay factor, and $\nabla_{\mathbf{V}}^{(t)}$ represents the gradient at the current step. For geometric priors, we evaluate curvature $\mathcal{K}_f$, which geometrically assesses local complexity. We estimate $\mathcal{K}_f$ of face $\mathcal{F}$ by calculating the average angle between its normal $\mathbf{n}_f$ and the normals $\mathbf{n}_{adj}$ of all adjacent faces $\mathcal{F}_{adj}$:

\vspace{-1mm}
\begin{equation}
\mathcal{K}_f = \frac{1}{N_{adj}} \sum_{\mathcal{F}_{adj}} \arccos(\mathbf{n}_f \cdot \mathbf{n}_{adj})
\end{equation}

\noindent When performing splitting, we first calculate EMA of the gradient magnitude $\mathcal{G}_f$ for each face (i.e., the mean $\mathcal{G}_v$ of three vertices). To avoid ineffectual splitting in already detailed regions, candidate faces are limited to the top 50\% by area. A weighted score $S_f$ combining optimization needs and geometric complexity is used for sampling:

\vspace{-2mm}
\begin{equation}
S_f = \alpha \mathcal{G}_{f} + \beta \mathcal{K}_{f}
\end{equation}

\noindent where $\alpha$ and $\beta$ are hyperparameters to balance the weights of gradient and curvature. The sampled faces are selected as splitting targets for the current round.

\noindent\textbf{Splitting Operation.} For a selected face $\mathcal{F}(v_a, v_b, v_c)$, we select the edge with the highest score, computed as the ratio of edge length to vertex degree, i.e., $S_e = l_e \mathbin{/} d_e$, as the splitting edge, which jointly considers geometric and topological factors. The main challenge is to determine the position of new vertex $v_s$ (see Figure~\ref{fig:vertex_split_merge} (a)). If $e$ is a non-boundary edge, it must be shared with another face $\mathcal{F}'(v_a, v_b, v_d)$. To fit local surface morphology, we project $v_c$ and $v_d$ onto $e$ and set $v_s = (t_c + t_d) / 2$, where $t_c$ and $t_d$ are projection points. Subsequently, topology update is performed by deleting two original faces $\mathcal{F}$ and $\mathcal{F}'$, then adding four new faces $\mathcal{F}(v_a, v_s, v_c)$, $\mathcal{F}(v_s, v_b, v_c)$, $\mathcal{F}(v_a, v_s, v_d)$, and $\mathcal{F}(v_s, v_b, v_d)$. Conversely, if $e$ is a boundary edge, adjacent face $\mathcal{F}'$ does not exist. In this case, we place $v_s$ at the midpoint of $e$, then delete $\mathcal{F}$, add two new faces $\mathcal{F}(v_a, v_s, v_c)$ and $\mathcal{F}(v_s, v_b, v_c)$.

Finally, to ensure the stability of topology update, two constraints are applied. The current split operation will be skipped if: (1) the adjacent face $\mathcal{F}'$ has already been modified in previous operations, to avoid concurrent modification conflicts; or (2) the relative position $\alpha = ||v_s - v_a|| / ||v_b - v_a||$ of $v_s$ on edge $e$ must fall within the central interval $[0.25, 0.75]$, to avoid introducing degenerate faces.

\begin{figure}[t]
    \centering
    \includegraphics[width=0.95\linewidth]{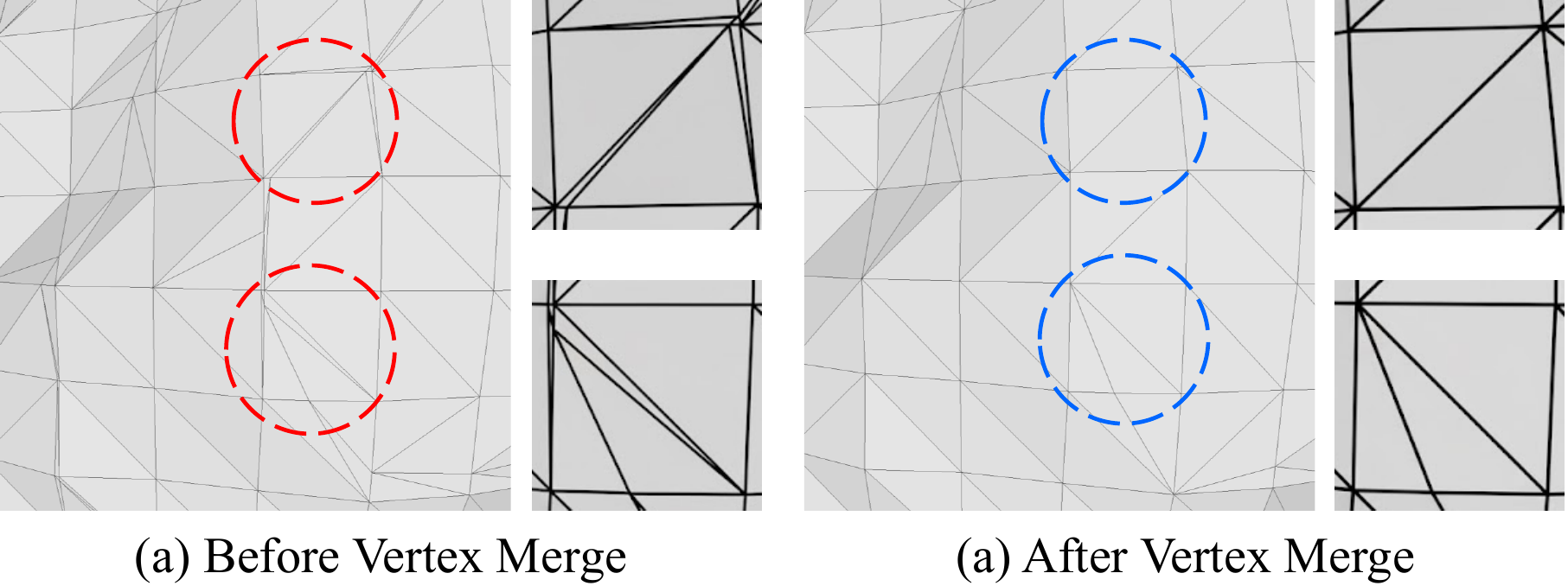}
    \vspace{-2mm}
    \caption{Visualization of our vertex merge operation, which eliminates degenerate faces to restore a clean topology.}
    \label{fig:method_merge}
    \vspace{-4mm}
\end{figure}

\begin{figure*}[t]
    \centering
    \includegraphics[width=\linewidth]{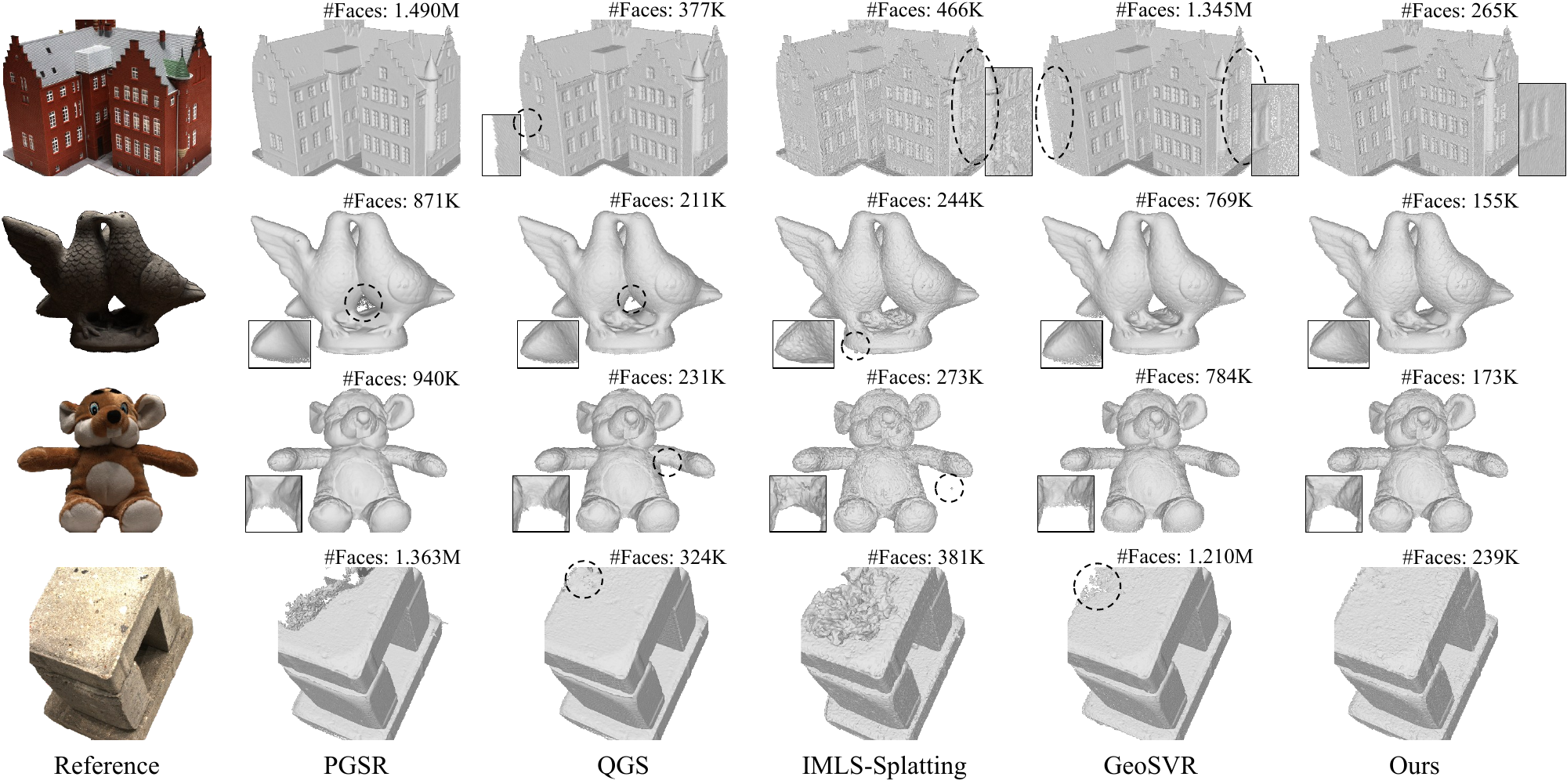}
    \vspace{-5mm}
    \caption{Qualitative geometric comparison on the DTU dataset. Our method produces structurally clean and high-fidelity meshes.}
    \label{fig:DTU1}
    \vspace{-3mm}
\end{figure*}

\subsection{Adaptive Vertex Merging}
\label{sec:merge}
As the counterpart to vertex splitting, the vertex merging mechanism is analogous to the classic ``edge collapse" in mesh simplification \cite{hoppe1996progressive,garland1997qem}. It aims to restore clean topology by removing redundant and degenerate faces.

\noindent\textbf{Merging Criterion.} Whether a face $\mathcal{F}$ needs to be merged is judged from two perspectives: rendering visibility and geometric morphology. Rendering visibility is measured by tracking a render contribution count $\mathcal{C}_{render}(\mathcal{F})$, which records how many times each face $\mathcal{F}$ contributes to the rendered output after rasterization. If $\mathcal{C}_{render}(\mathcal{F}) = 0$ for a face, it indicates the face was never rendered during training (e.g., becoming an internal face or being completely occluded), and is thus considered redundant. For geometric morphology, to address the degenerate faces produced during optimization, we calculate each face's degeneracy $\mathcal{D}_f$:

\vspace{-2mm}
\begin{equation}
\mathcal{D}_f = \frac{\text{Area}(\mathcal{F})}{l_{\max}^2(\mathcal{F})}
\end{equation}

\noindent where $l_{\max}(\mathcal{F})$ denotes the length of the longest edge in face $\mathcal{F}$. If $\mathcal{D}_f$ is less than threshold $\tau_{degen}$, the face is considered degenerate. Finally, only faces in the bottom 50\% by area that meet either criterion are added to the merge set $\mathcal{M}_{merge}$,  restricting the process to fine local details. Figure~\ref{fig:method_merge} intuitively demonstrates this process; our mechanism can identify these degenerate forms and clean them up.

\noindent\textbf{Merging Operation.} For each face $\mathcal{F}(v_i, v_j, v_m)$ in $\mathcal{M}_{merge}$, we first determine the edge $e_{collapse}$ to collapse. If $\mathcal{F}$ is a boundary face, we collapse its boundary edge $e_{collapse} = (v_i, v_m)$ to maintain boundary integrity. If $\mathcal{F}$ is an interior face, we collapse edge $e_{collapse}$ with the smallest $S_e$, in order to efficiently simplify the local mesh structure. After selecting $e_{collapse}$, we compare the vertex degrees of its two endpoints, $v_i$ and $v_m$. If $v_m$ has a lower degree, it will be merged into $v_i$.

The topology update operation is shown in Figure~\ref{fig:vertex_split_merge} (b): first, vertex $v_m$ is deleted. For adjacent faces that contained both $v_i$ and $v_m$ (i.e., $\mathcal{F}(v_i, v_j, v_m)$ and $\mathcal{F}(v_i, v_k, v_m)$), they will be degenerated into an edge, so we directly delete them. For adjacent faces that contained $v_m$ but not $v_i$ (e.g., $\mathcal{F}(v_j, v_l, v_m)$ and $\mathcal{F}(v_l, v_k, v_m)$), they must be re-linked to preserve the local manifold structure. We replace $v_m$ with $v_i$ in these faces, updating them to $\mathcal{F}(v_j, v_l, v_i)$ and $\mathcal{F}(v_l, v_k, v_i)$. Thus, each merge operation in the non-boundary case deterministically deletes the two faces and re-links the remaining adjacent faces of $v_m$. To avoid topological errors from concurrent modifications, a merge operation will be skipped if a face requiring modification has already been affected by previous operations.

\begin{table*}[t]
    \centering
    \caption{Quantitative geometric comparison on the DTU dataset. Our method achieves comparable accuracy to SOTA methods while using significantly fewer faces. {\color{colorfirst}\raisebox{-0.15mm}{\rule{3mm}{2mm}}} denotes the best, {\color{colorsecond}\raisebox{-0.15mm}{\rule{3mm}{2mm}}} second best, {\color{colorthird}\raisebox{-0.15mm}{\rule{3mm}{2mm}}} third best, respectively.}
    \label{tab:dtu_comparison}
    
    \resizebox{\textwidth}{!}{
    \begin{tabular}{ll cccccccccccccccc c c}
        \toprule
        & Method & 24 & 37 & 40 & 55 & 63 & 65 & 69 & 83 & 97 & 105 & 106 & 110 & 114 & 118 & 122 & Avg. & Time & Faces \\
        \midrule
        
        \multirow{4}{*}{\shortstack{NeRF-\\based}} 
        & VolSDF~\cite{yariv2021volsdf} & 1.14 & 1.26 & 0.81 & 0.49 & 1.25 & 0.70 & 0.72 & 1.29 & 1.18 & 0.70 & 0.66 & 1.08 & 0.42 & 0.61 & 0.55 & 0.86 & $>$12h & 1.58M \\
        & NeuS~\cite{wang2021neus} & 0.83 & 0.98 & 0.56 & 0.37 & 1.13 & 0.59 & 0.60 & 1.45 & 0.95 & 0.78 & 0.52 & 1.43 & 0.36 & 0.45 & 0.45 & 0.77 & $>$12h & 798K \\
        & Neuralangelo~\cite{tancik2023neuralangelo} & 0.45 & 0.74 & 0.33 & \cellcolor{colorthird}{0.34} & 1.05 & \cellcolor{colorsecond}{0.54} & 0.53 & 1.33 & 1.05 & 0.72 & \cellcolor{colorsecond}{0.43} & 0.69 & 0.34 & \cellcolor{colorthird}{0.38} & 0.42 & 0.62 & $>$12h & 1.98M \\
        \midrule
        
        \multirow{7}{*}{\shortstack{GS-\\based}} 
        & SuGaR~\cite{su2023sugar} & 1.47 & 1.33 & 1.13 & 0.61 & 2.25 & 1.71 & 1.15 & 1.63 & 1.62 & 1.07 & 0.79 & 2.45 & 0.98 & 0.88 & 0.79 & 1.33 & 1h & 962K \\
        & 2DGS ~\cite{huang2024twodgs} & 0.46 & 0.84 & \cellcolor{colorthird}{0.31} & 0.45 & 0.92 & 1.01 & 0.83 & 1.23 & 1.30 & 0.66 & 0.61 & 1.07 & 0.45 & 0.71 & 0.54 & 0.76 & \cellcolor{colorfirst}{11m} & 260K \\
        & TriSplat~\cite{held2025trianglesplatting} & 0.76 & 0.97 & 0.77 & 0.46 & 1.77 & 1.27 & 1.04 & 1.75 & 1.72 & 0.80 & 0.91 & 1.20 & 0.56 & 0.91 & 0.97 & 1.05 & \cellcolor{colorthird}{15m} & 265K \\
        & GOF ~\cite{yu2024gof} & 0.50 & 0.82 & 0.37 & 0.37 & 1.12 & 0.74 & 0.73 & 1.18 & 1.29 & 0.68 & 0.77 & 0.90 & 0.42 & 0.66 & 0.49 & 0.74 & 1h & 1.06M \\
        & PGSR ~\cite{chen2024pgsr} & \cellcolor{colorsecond}{0.36} & \cellcolor{colorsecond}{0.57} & 0.38 & \cellcolor{colorfirst}{0.33} & 0.78 & 0.58 & \cellcolor{colorsecond}{0.50} & \cellcolor{colorthird}{1.08} & \cellcolor{colorsecond}{0.63} & \cellcolor{colorthird}{0.59} & 0.46 & \cellcolor{colorsecond}{0.54} & \cellcolor{colorfirst}{0.30} & \cellcolor{colorthird}{0.38} & \cellcolor{colorsecond}{0.34} & \cellcolor{colorsecond}{0.52} & 30m & 1.05M \\
        & QGS ~\cite{zhang2025qgs} & \cellcolor{colorthird}{0.38} & \cellcolor{colorthird}{0.62} & 0.37 & 0.38 & \cellcolor{colorsecond}{0.75} & \cellcolor{colorthird}{0.55} & \cellcolor{colorthird}{0.51} & 1.12 & \cellcolor{colorthird}{0.68} & 0.61 & 0.46 & \cellcolor{colorthird}{0.58} & 0.35 & 0.41 & 0.40 & \cellcolor{colorthird}{0.54} & 48m & 252K \\
        \midrule
        
        \multirow{5}{*}{\shortstack{Mesh-\\driven}} 
        & Nvdiffrec~\cite{munkberg2022nvdiffrec} & 3.04 & 3.02 & 2.10 & 0.78 & 2.18 & 1.60 & 1.46 & 1.67 & 2.85 & 1.26 & 1.10 & 3.26 & 1.13 & 1.31 & 1.19 & 1.86 & $>$1h & \cellcolor{colorfirst}{132K} \\
        & Flexicubes~\cite{shen2023flexicubes} & 1.66 & 1.60 & 0.91 & 0.50 & 2.60 & 1.32 & 0.87 & 1.45 & 1.60 & 1.38 & 0.73 & 1.85 & 1.01 & 0.64 & 0.75 & 1.26 & $>$1h & \cellcolor{colorsecond}{168K}\\
        & IMLS-Splat~\cite{yang2025imlssplatting} & 0.63 & 1.21 & 0.67 & 0.62 & 1.04 & 0.82 & 0.78 & 1.32 & 1.09 & 0.89 & 0.69 & 0.95 & 0.62 & 0.63 & 0.65 & 0.84 & \cellcolor{colorthird}{15m} & 418K \\
        & GeoSVR ~\cite{li2025geosvr} & \cellcolor{colorfirst}{0.32} & \cellcolor{colorfirst}{0.51} & \cellcolor{colorfirst}{0.30} & \cellcolor{colorfirst}{0.33} & \cellcolor{colorfirst}{0.71} & \cellcolor{colorfirst}{0.48} & \cellcolor{colorfirst}{0.42} & \cellcolor{colorfirst}{1.03} & \cellcolor{colorfirst}{0.62} & \cellcolor{colorfirst}{0.56} & \cellcolor{colorfirst}{0.33} & \cellcolor{colorfirst}{0.46} & \cellcolor{colorfirst}{0.30} & \cellcolor{colorfirst}{0.34} & \cellcolor{colorfirst}{0.32} & \cellcolor{colorfirst}{0.47} & 49m & 1.12M \\
        & Ours & 0.44 & 0.70 & \cellcolor{colorfirst}{0.30} & \cellcolor{colorthird}{0.34} & \cellcolor{colorthird}{0.76} & 0.63 & 0.52 & \cellcolor{colorsecond}{1.06} & 1.05 & \cellcolor{colorsecond}{0.58} & \cellcolor{colorthird}{0.45} & 0.88 & \cellcolor{colorthird}{0.32} & \cellcolor{colorsecond}{0.36} & \cellcolor{colorthird}{0.38} & 0.58 & \cellcolor{colorsecond}{13m} & \cellcolor{colorthird}{196K} \\
        \bottomrule
    \end{tabular}
    }
	\vspace{-2mm}
\end{table*}

\subsection{Geometry and Texture Optimization}
Our framework adopts a decoupled geometry and texture optimization strategy, where geometry is represented by the vertex positions $\mathbf{V}$ of mesh $\mathcal{M}$, while texture is stored in a separate, fixed-resolution UV map $\mathbf{T}$, with UV coordinates $\mathbf{u}$ linking them. In the rendering pipeline, vertex positions $\mathbf{V}$ determine the mesh projection, while interpolated UV coordinates $\mathbf{u}$ are used to sample the UV map $\mathbf{T}$ for surface color~\cite{munkberg2022nvdiffrec}. Image loss gradients are independently back-propagated to drive geometry and texture optimization.

\noindent\textbf{Geometry Optimization.} Geometry optimization updates all vertex positions $\mathbf{V}$ to align the rendered mesh shape with the GT image. This is achieved by minimizing a composite loss function (see Sec.~\ref{sec:training_strategy}), which provides gradients that push $\mathbf{V}$ toward the correct spatial positions when discrepancies exist between the rendered and GT images.

\noindent\textbf{Texture Optimization.} Texture optimization adjusts all texels on the UV map $\mathbf{T}$ by back-propagating image rendering loss through the differentiable UV sampling operation, which effectively ``paints" high-definition texture onto the mesh. This decoupling allows the resolution of $\mathbf{T}$ to be independent of the face count, enabling our method to represent high-resolution texture details with a concise mesh and thus avoid face count inflation.

The texture decoupling mechanism must also coordinate with dynamic topology updates to correctly maintain UV coordinates. When a vertex split operation creates a new vertex $v_s$, its UV coordinate $u_s$ should be initialized appropriately. If $v_s$ is on a boundary edge $e(v_a, v_b)$, its UV coordinate $u_s$ is set to the midpoint of $u_a$ and $u_b$. For non-boundary edges, $v_s$ is shared by faces $\mathcal{F}(v_a, v_b, v_c)$ and $\mathcal{F}'(v_a, v_b, v_d)$, which may lie on different UV islands. We perform interpolation in the UV space of $\mathcal{F}$ and $\mathcal{F}'$ to obtain two candidate UV coordinates: $u_s^{(1)} = (1-\mu)u_a + \mu u_b$ and $u_s^{(2)} = (1-\mu)u_a' + \mu u_b'$, where $u_a'$ and $u_b'$ are the UV coordinates of $v_a$ and $v_b$ in $\mathcal{F}'$, and $\mu$ is the relative position of $v_s$ on edge $e$. If $\lVert u_s^{(1)} - u_s^{(2)} \rVert < \tau_{uv}$, then set $u_s$ to their average, where $\tau_{uv}$ is the threshold. Conversely, $v_s$ is associated with two distinct UV coordinates, $u_s^{(1)}$ and $u_s^{(2)}$, which are stored independently to ensure a proper texture seam. Correspondingly, when a vertex $v_m$ is removed in vertex merging, any unreferenced UV coordinates are deleted.

\subsection{Training Strategy}
\label{sec:training_strategy}
Our model is trained using a composite loss $\mathcal{L}$ defined as:

\vspace{-4mm}
\begin{equation}
\mathcal{L} = \lambda_{rgb}\mathcal{L}_{rgb} + \lambda_{d}\mathcal{L}_{d} + \lambda_{m}\mathcal{L}_{m} + \lambda_{s}\mathcal{L}_{s} + \lambda_{b}\mathcal{L}_{b}
\end{equation}

\noindent where $\lambda$ is the weight coefficient for each term. Specifically, $\mathcal{L}_{rgb}$ is the photometric loss, using the same $(1 - \lambda_{dssim})\mathcal{L}_{L1} + \lambda_{dssim}\mathcal{L}_{D-SSIM}$ combination as 3DGS \cite{kerbl2023gaussian}. $\mathcal{L}_{d}$ is the Pearson depth loss, which leverages depth reference from Depth Anything 3~\cite{lin2025depthanything3}. While $\mathcal{L}_{m}$ is the silhouette loss, defined as the binary cross-entropy between predicted and ground-truth masks~\cite{liu2019softras,kato2018neural}. To ensure geometric smoothness, $\mathcal{L}_{s}$ applies Laplacian smoothing~\cite{desbrun1999implicit,taubin1995signal}. Finally, $\mathcal{L}_{b}$ is the bi-vertex offset regularization loss as defined in FlexiCubes \cite{shen2023flexicubes}.

\section{Experiments}
\label{sec:experiments}

\begin{figure}[t]
    \centering
    \includegraphics[width=\linewidth]{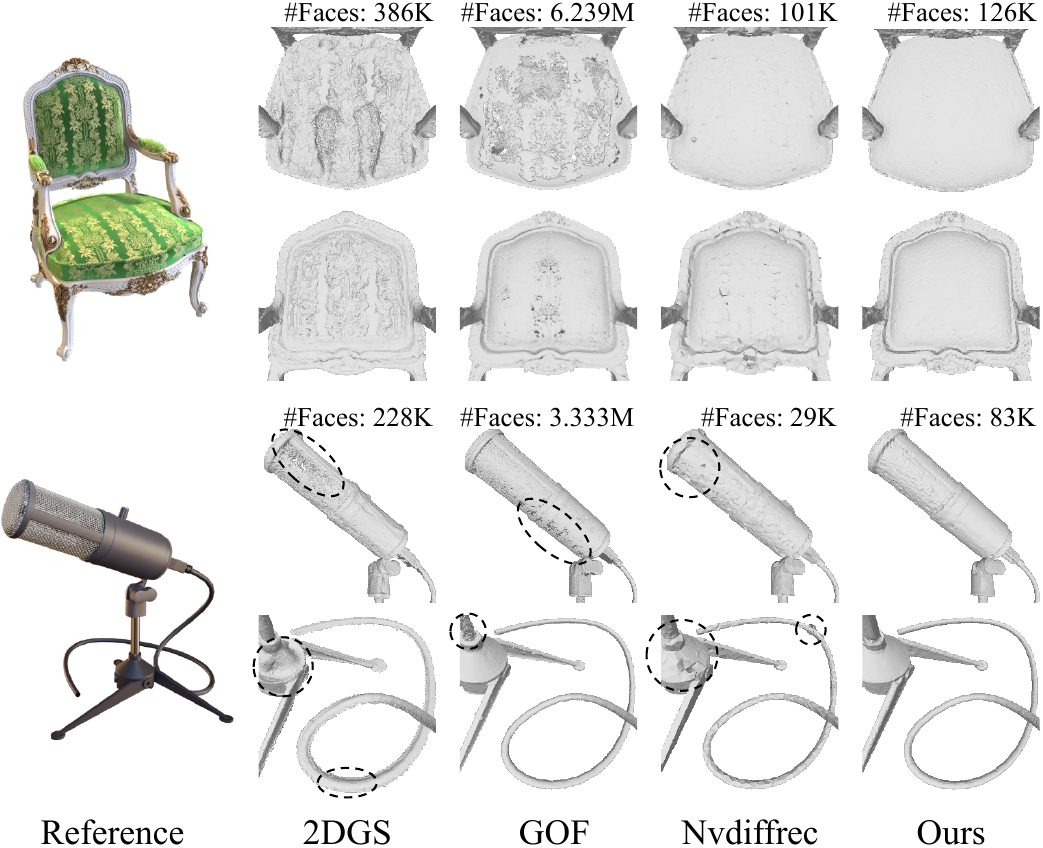}
    \vspace{-5mm}
    \caption{Qualitative comparison on the NeRF-synthetic dataset.}
    \label{fig:nerf1}
    \vspace{-4mm}
\end{figure}

\subsection{Implementation Details}
\label{sec:implementation_details}
Our method is implemented based on the PyTorch framework, utilizing CUDA kernels for acceleration, and using nvdiffrast \cite{laine2020modular} for differentiable rendering. All experiments were completed on a single NVIDIA RTX 3090 GPU. Our pipeline begins with an initial coarse mesh, which is obtained by training 2DGS \cite{huang2024twodgs} for 7k iterations (about 2 mins) and extracting it using TSDF \cite{curless1996volumetric} at a resolution of $256^3$. Before training, we perform multiple rounds of vertex merging to eliminate degenerate faces in the initial mesh. During optimization, firstly, a 1k-iteration warm-up is used for initial geometry and texture fitting. Then, during the 1k–7k phase, vertex splitting and merging are performed every 500 iterations and the UV coordinates are updated accordingly, with UV map reconstruction every 2000 iterations \cite{levy2002lscm, sheffer2005abfpp}. Finally, in the last 1k iterations, the topology is frozen for final refinement. Detailed hyperparameters and loss weights are provided in the supplementary material.

\subsection{Comparison}
\textbf{Datasets.} Our experiments are conducted on two benchmark datasets. The DTU dataset \cite{jensen2014dtu} contains 15 real-world objects, each including 49 or 69 images at a resolution of $1600 \times 1200$. For efficiency, we downsample the images to $800 \times 600$ resolution, as in~\cite{wang2021neus,yariv2021volsdf}. NeRF-synthetic dataset \cite{mildenhall2020nerf} contains 8 object-level scenes, each with 100 training images and 200 test images at $800 \times 800$ resolution.

\begin{table}[t]
    \centering
    \caption{Quantitative comparison on the NeRF-synthetic dataset.}
    \label{tab:nerf_comparison}
    \vspace{-2mm}
    \resizebox{\columnwidth}{!}{
    \begin{tabular}{l ccccccccc c c}
        \toprule
        Method & Mic & Chair & Ship & Material & Lego & Drums & Ficus & Hotdog & Avg. & Time & Faces \\
        \midrule
        
        NeuS~\cite{wang2021neus} & 0.53 & \cellcolor{colorthird}{0.37} & 1.25 & \cellcolor{colorfirst}{0.35} & 0.83 & \cellcolor{colorsecond}{0.86} & \cellcolor{colorsecond}{0.40} & 0.96 & \cellcolor{colorthird}{0.69} & $>$12h & \\
        Neuralangelo~\cite{tancik2023neuralangelo} & \cellcolor{colorthird}{0.52} & \cellcolor{colorfirst}{0.29} & \cellcolor{colorfirst}{0.34} & 0.85 & \cellcolor{colorfirst}{0.61} & \cellcolor{colorfirst}{0.77} & 0.54 & 0.87 & \cellcolor{colorfirst}{0.60} & $>$12h & 8.387M\\
        2DGS\textdagger ~\cite{huang2024twodgs} & 1.39 & 0.92 & 1.92 & 1.11 & 0.75 & 0.88 & 1.35 & 0.92 & 1.16 & \cellcolor{colorfirst}{11m} & 639K \\
        TriSplat & 1.14 & 1.65 & 2.17 & 1.15 & 0.70 & 1.35 & 2.72 & 1.03 & 1.49 & \cellcolor{colorthird}{15m} & 623K \\
        GOF\textdagger ~\cite{yu2024gof} & 0.81 & 1.08 & \cellcolor{colorsecond}{0.91} & 0.83 & 0.78 & 0.89 & \cellcolor{colorthird}{0.52} & \cellcolor{colorthird}{0.70} & 0.82 & 1h & 7.32M \\
        Nvdiffrec~\cite{munkberg2022nvdiffrec} & \cellcolor{colorthird}{0.52} & 0.45 & 2.48 & 0.73 & 0.71 & 1.33 & 0.61 & 0.69 & 0.94 & $>$1h & \cellcolor{colorfirst}{81K} \\
        Flexicubes~\cite{shen2023flexicubes} & \cellcolor{colorsecond}{0.46} & 0.44 & 5.23 & \cellcolor{colorthird}{0.55} & \cellcolor{colorthird}{0.70} & 1.05 & \cellcolor{colorfirst}{0.26} & \cellcolor{colorsecond}{0.68} & 1.17 & $>$1h & \cellcolor{colorsecond}{137K}\\
        IMLS-Splat~\cite{yang2025imlssplatting} & 0.64 & 0.49 & \cellcolor{colorthird}{1.19} & 0.59 & 0.74 & \cellcolor{colorsecond}{0.86} & 0.56 & 0.74 & 0.73 & \cellcolor{colorthird}{15m} & 578K \\
        Ours & \cellcolor{colorfirst}{0.44} & \cellcolor{colorsecond}{0.31} & 1.24 & \cellcolor{colorsecond}{0.43} & \cellcolor{colorsecond}{0.66} & \cellcolor{colorthird}{0.91} & 0.56 & \cellcolor{colorfirst}{0.61} & \cellcolor{colorsecond}{0.64} & \cellcolor{colorsecond}{13m} & \cellcolor{colorthird}{216K} \\

        \bottomrule
    \end{tabular}
    }
    \vspace{-4mm}
\end{table}

\noindent\textbf{Baselines.} We compare our method against three categories of surface reconstruction methods: (1) NeRF-based methods, including VolSDF \cite{yariv2021volsdf}, NeuS \cite{wang2021neus}, and Neuralangelo \cite{tancik2023neuralangelo}. (2) GS-based methods, including SuGaR \cite{su2023sugar}, 2DGS \cite{huang2024twodgs}, Triangle Splatting \cite{held2025trianglesplatting}, GOF \cite{yu2024gof}, PGSR \cite{chen2024pgsr}, and QGS \cite{zhang2025qgs}. (3) Mesh-driven methods, including Nvdiffrec \cite{munkberg2022nvdiffrec}, FlexiCubes \cite{shen2023flexicubes}, IMLS-Splatting \cite{yang2025imlssplatting}, and GeoSVR \cite{li2025geosvr}. Our Nvdiffrec and FlexiCubes baselines utilize the Nvdiffrec framework with DMTet and FlexiCubes backends respectively. Note that our some methods \cite{zhang2025qgs,munkberg2022nvdiffrec,shen2023flexicubes,yang2025imlssplatting,li2025geosvr,tancik2023neuralangelo, wang2021neus, huang2024twodgs,held2025trianglesplatting} require or benefit from object masks on DTU. For fairness, masks are provided to those methods that benefit from them.

\begin{figure}[H]
    \centering
    \includegraphics[width=\linewidth]{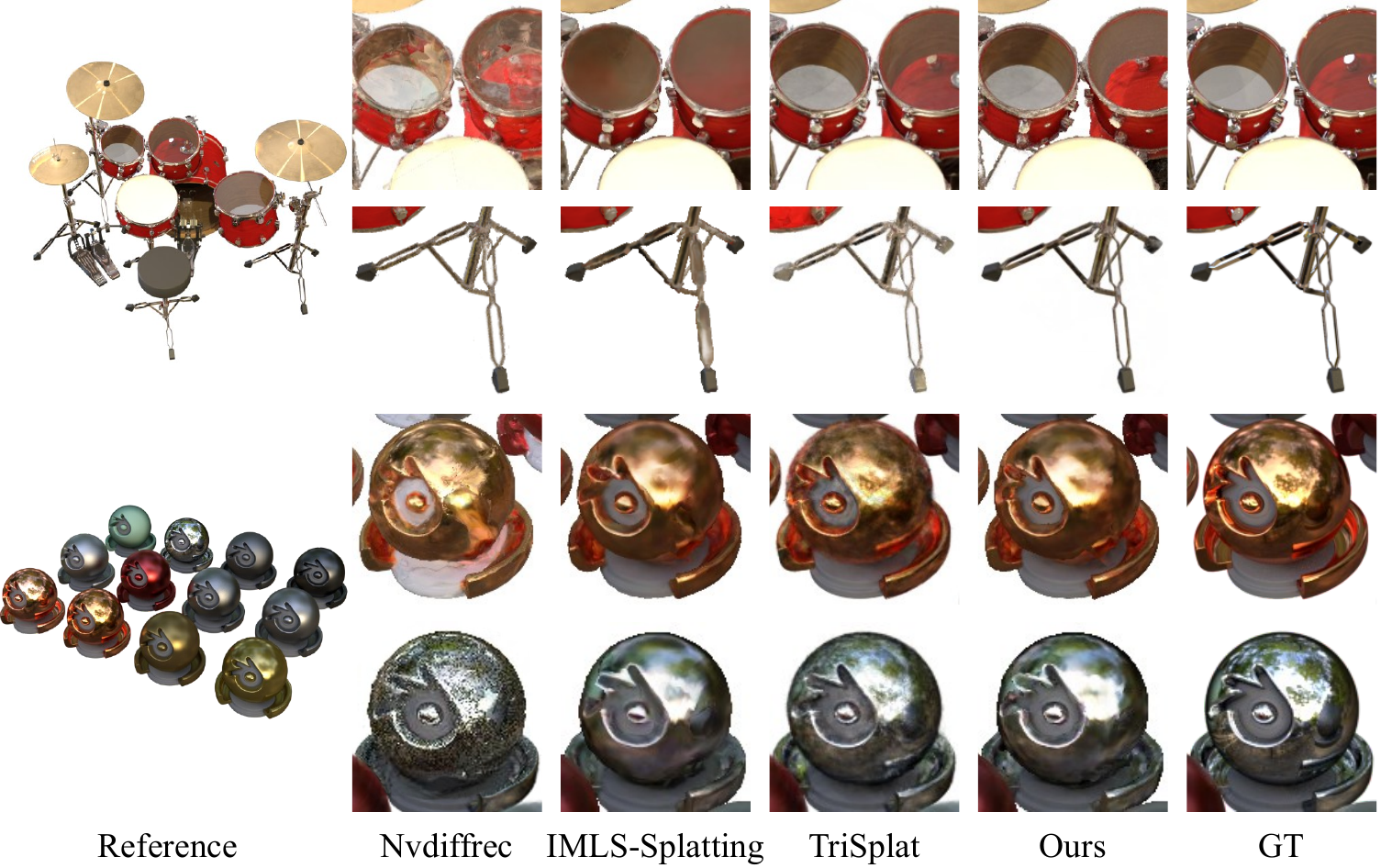}
    \vspace{-5mm}
    \caption{Qualitative rendering comparison on the NeRF-synthetic dataset. Compared to other mesh-driven methods, our method produces higher-fidelity results with more detailed textures.}
    \label{fig:nerf2}
    \vspace{-3mm}
\end{figure}

\noindent\textbf{Geometry Evaluation.} To evaluate the geometric quality of the reconstructed meshes, we use Chamfer Distance (CD) for quantitative comparison. Results on DTU dataset are shown in Table \ref{tab:dtu_comparison}. For our method, the reported runtime includes both initialization and training. Our method achieves geometric accuracy comparable to SOTA methods while demonstrating a significant advantage in efficiency, requiring fewer faces and shorter training time, which can also be observed in Figure~\ref{fig:DTU1}. GeoSVR \cite{li2025geosvr} and PGSR \cite{chen2024pgsr} generate meshes with a large number of redundant faces and visible floating artifacts. While meshes from QGS \cite{zhang2025qgs} and IMLS-Splatting \cite{yang2025imlssplatting} also show rough edges and suspended mesh fragments. In contrast, ExMesh produces meshes without obvious artifacts, recovering details while maintaining smooth and intact surfaces.

Compared to DTU dataset, the NeRF-synthetic dataset is more challenging due to its complex topology and diverse reflective properties. As shown in Table~\ref{tab:nerf_comparison}, our method achieves geometric accuracy comparable to state-of-the-art methods. Figure~\ref{fig:nerf1} shows the visual comparison on the NeRF-synthetic dataset. 2DGS \cite{huang2024twodgs} and GOF \cite{yu2024gof} tend to produce multi-layered suspended mesh artifacts and introduce significant noise when fitting slender structures. Meshes reconstructed by Nvdiffrec \cite{munkberg2022nvdiffrec} exhibit obvious polygonal faceting. In comparison, our method reconstructs smoother surfaces and captures fine engraved details that are not present in the results of other mesh-driven methods.

\begin{table}[t]
    \centering
    \caption{Rendering quality results on the NeRF-synthetic dataset.}
    \label{tab:nerf_rendering}
    \vspace{-3mm}
    \resizebox{0.88\columnwidth}{!}{
    \begin{tabular}{ll ccc}
        \toprule
         & Method & PSNR & SSIM & LPIPS \\
        \midrule
        
        \multirow{4}{*}{\shortstack{GS-\\based}} 
        & 3DGS~\cite{kerbl2023gaussian} & 33.32 & 0.969 & 0.030 \\
        & 2DGS~\cite{huang2024twodgs} & 33.07 & 0.968 & 0.031 \\
        & TriSplat~\cite{held2025trianglesplatting} & 33.87 & 0.97 & 0.03 \\
        \midrule

        \multirow{4}{*}{\shortstack{Mesh-\\driven}} 
        & Nvdiffrec~\cite{munkberg2022nvdiffrec} & 26.87 & 0.930 & 0.090 \\
        & Flexicubes~\cite{shen2023flexicubes} & 27.50 & 0.930 & 0.080 \\
        & IMLS-Splat~\cite{yang2025imlssplatting} & 28.38 & 0.950 & 0.060 \\
        & Ours & 29.32 & 0.958 & 0.051 \\
        
        \bottomrule
    \end{tabular}
    }
    \vspace{-5mm}
\end{table}

\noindent\textbf{Rendering Evaluation.} We further evaluate the novel view synthesis quality on the NeRF-synthetic dataset. As shown in Table \ref{tab:nerf_rendering}, our method outperforms other mesh-driven methods on all rendering metrics (PSNR, SSIM \cite{wang2004ssim}, LPIPS \cite{zhang2018lpips}). The qualitative comparison in Figure~\ref{fig:nerf2} also shows that, compared to Nvdiffrec \cite{munkberg2022nvdiffrec} and IMLS-Splatting \cite{yang2025imlssplatting}, our rendered results exhibit higher fidelity and more detailed textures. However, there is still a gap in rendering quality when compared to Gaussian-based methods, as they leverage Spherical Harmonics (SH) to flexibly capture complex view-dependent appearance. Bridging this rendering gap for explicit meshes remains an ongoing challenge.

\vspace{-1mm}
\begin{figure}[H]
    \centering
    \includegraphics[width=0.95\linewidth]{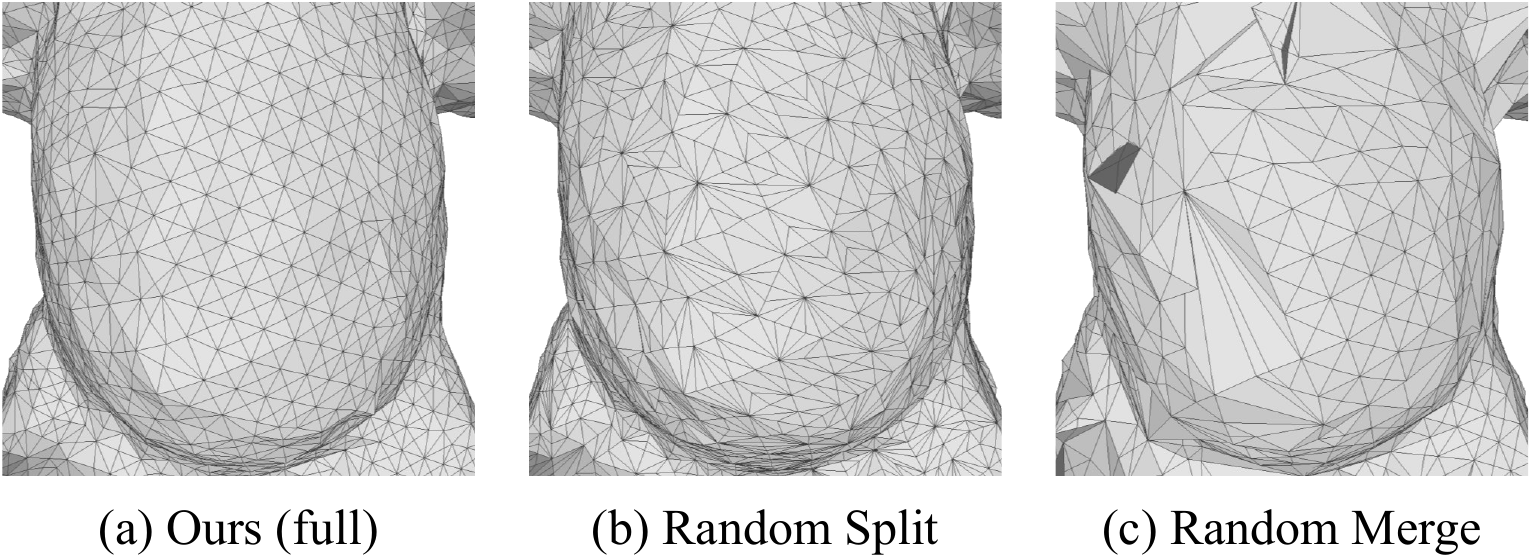}
    \vspace{-2mm}
    \caption{Ablation study on our vertex split and merge strategies.}
    \label{fig:ablation_split_merge}
    \vspace{-3mm}
\end{figure}

\subsection{Ablation Study}

\noindent\textbf{Vertex Splitting and Merging.} We first validate effectiveness of the dynamic topology adjustment by comparing five variants: (1) only merge; (2) only split; (3) random split, where the edge to split is selected by our criterion but the split position is set to a random point; (4) random merge; and (5) our full method. Quantitative results on the NeRF-synthetic dataset are shown in Table \ref{tab:ablation_split_merge}. "‘Only Merge’ fails to add faces for detailed fitting,. The ``Only Split" variant, while able to add details, led to a substantial increase in face count. Random strategies also perform worse than our full method. Qualitative comparison in Figure~\ref{fig:ablation_split_merge} intuitively reveals the reasons. Random split disrupts the local regularity of the mesh, leading to an uneven and chaotic triangulation. Additionally, random merge incorrectly collapses edges, introducing severe topological irregularities.

\begin{figure}[t]
    \centering
    \includegraphics[width=\linewidth]{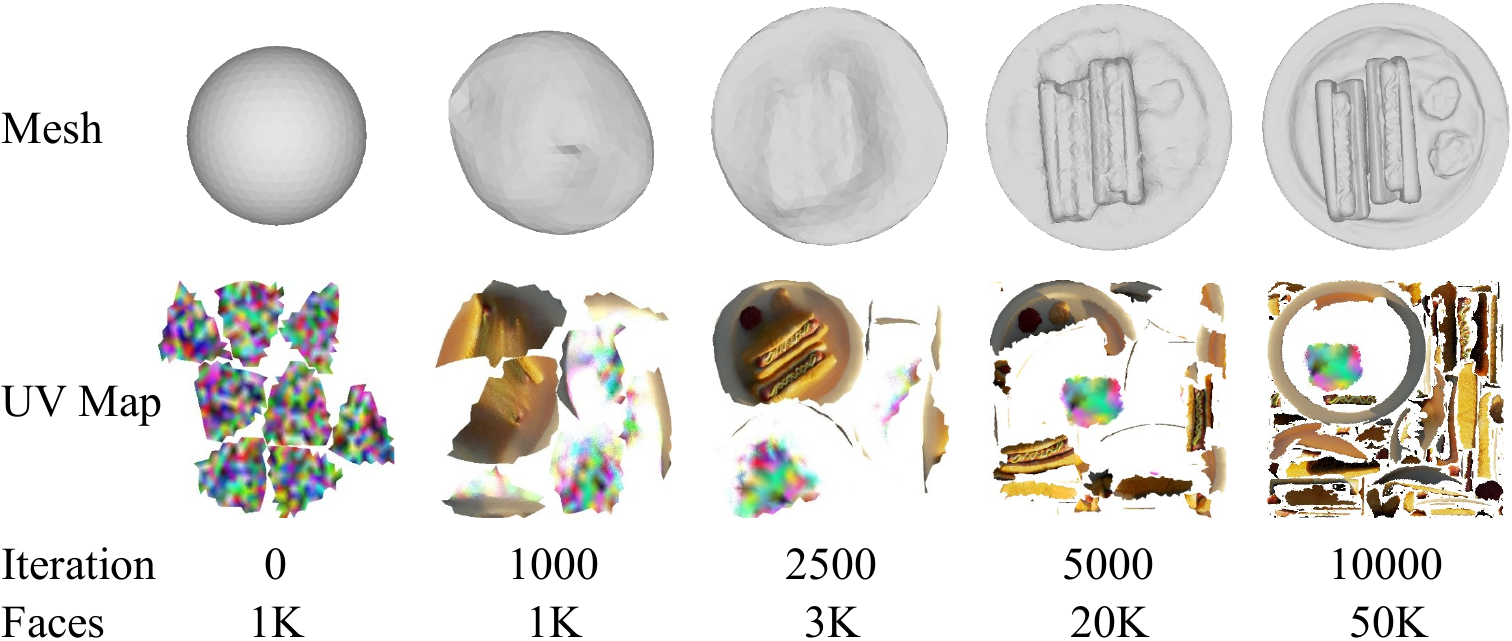}
    \vspace{-5mm}
    \caption{Ablation study on the initialization strategy.}
    \label{fig:ablation_wo_input}
    \vspace{-5mm}
\end{figure}

\noindent\textbf{Initialization Strategy.} Instead of using the initial coarse mesh described in Section \ref{sec:implementation_details}, we use a spherical mesh with 1000 faces and random colors as the initial input. As shown in Figure~\ref{fig:ablation_wo_input}, our full optimization pipeline can still deform and subdivide the mesh, eventually reconstructing the target object. We observe that, compared to our standard initialization strategy, training from an unrelated sphere requires more iterations to converge, and the final geometric accuracy is slightly lower. These results demonstrate that our method does not strictly depend on the initial mesh.

\begin{table}[t]
    \centering
    \caption{Quantitative results for the ablation study on our vertex splitting and merging strategies on the NeRF-synthetic dataset.}
    \vspace{-2mm}
    \label{tab:ablation_split_merge}
    \setlength{\tabcolsep}{6pt} 
    \renewcommand{\arraystretch}{1} 
    \resizebox{0.85\columnwidth}{!}{
    \begin{tabular}{lccc}
        \toprule
        Setting & CD$\downarrow$ & PSNR$\uparrow$ & Face Count \\
        \midrule
        Only Split       & 0.74 & 28.77 & 239K \\
        Only Merge       & 1.81 & 23.51 & 12K \\
        Random Split     & 0.77 & 28.30 & 184K \\
        Random Merge     & 1.34 & 26.27 & 192K \\
        Ours (full)      & 0.64 & 29.32 & 196K \\
        \bottomrule
    \end{tabular}
    }
    \vspace{-1mm}
\end{table}

\vspace{-1mm}
\begin{table}[h]
    \centering
    \caption{Ablation study on texture representation.}
    \vspace{-2mm}
    \label{tab:ablation_uvmap}
    \setlength{\tabcolsep}{6pt} 
    \renewcommand{\arraystretch}{1} 
    \resizebox{0.85\columnwidth}{!}{
    \begin{tabular}{lccc}
        \toprule
        Setting & CD$\downarrow$ & PSNR$\uparrow$ & Face Count \\
        \midrule
        Per-Vertex Color   & 0.76 & 28.47 & 356K \\
        Per-Face Color     & 0.78 & 28.12 & 442K  \\
        MLP Texture        & 0.64 & 28.87 & 198K \\
        Ours (UV Map)      & 0.64 & 29.32 & 196K \\
        \bottomrule
    \end{tabular}
    }
    \vspace{-2mm}
\end{table}

\noindent\textbf{UV Mapping.} Moreover, our standard method is evaluated alongside three alternatives: (1) Per-Vertex Color, which stores color directly on the vertices; (2) Per-Face Color; and (3) Coordinate-based MLP textures. As shown in Table \ref{tab:ablation_uvmap}, binding texture to geometry (per-vertex/face) requires a significantly denser mesh to capture high-frequency details and match our geometric accuracy. In contrast, our decoupled strategy achieves similar fidelity with far fewer faces. Furthermore, compared to coordinate-based MLP textures, our UV-based method yields comparable quality while maintaining an editable texture map throughout optimization, avoiding post-training baking.

\begin{figure}[t]
    \centering
    \includegraphics[width=0.9\linewidth]{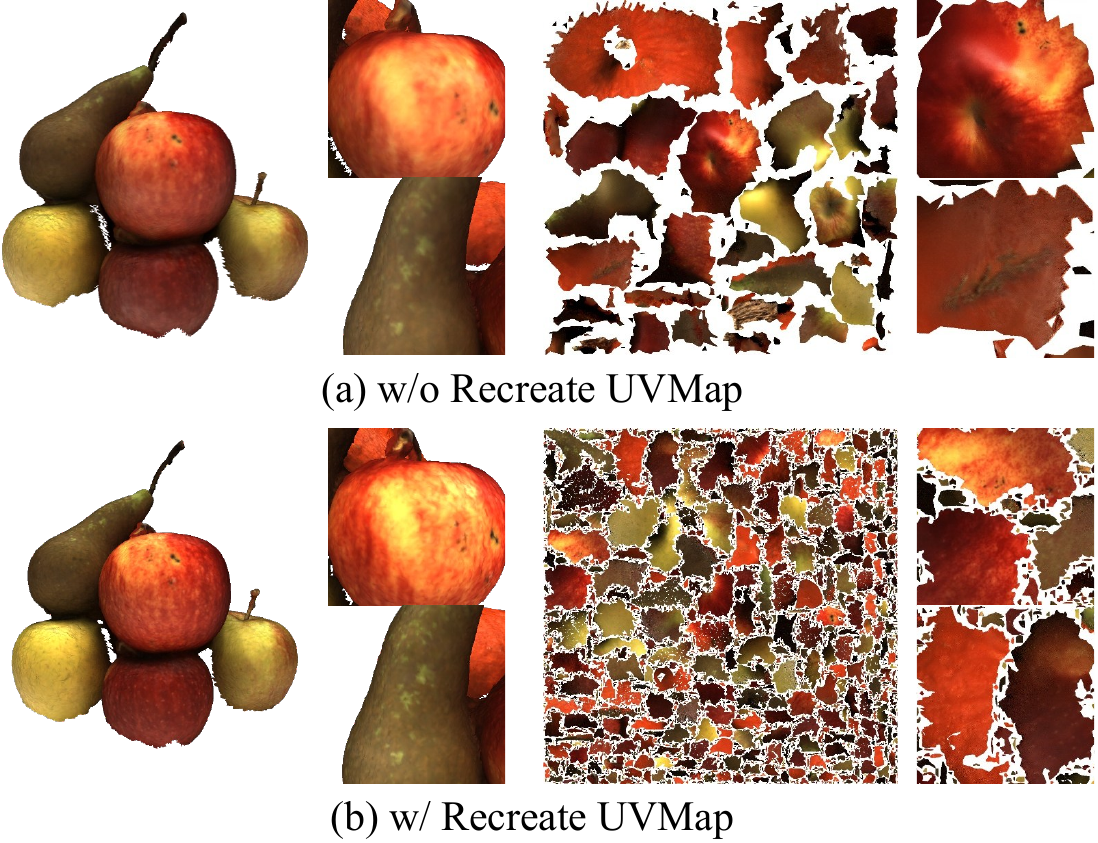}
    \vspace{-2mm}
    \caption{Ablation study on UV reconstruction.}
    \label{fig:ablation_recreate_uvmap}
    \vspace{-2mm}
\end{figure}

\noindent\textbf{UV Reconstruction.} We further evaluate the necessity of periodic UV map reconstruction by comparing our standard strategy with creating UV map only once at the beginning. As shown in Figure~\ref{fig:ablation_recreate_uvmap}, initial packing efficiency of UV islands is low. As the topology changes during training, a fixed UV map becomes inconsistent with the mesh, leading to texture stretching and UV island fragmentation. This forces fine details into small or distorted UV regions, limiting texture resolution and causing blurry results.

\begin{figure}[t]
    \centering
    \includegraphics[width=\linewidth]{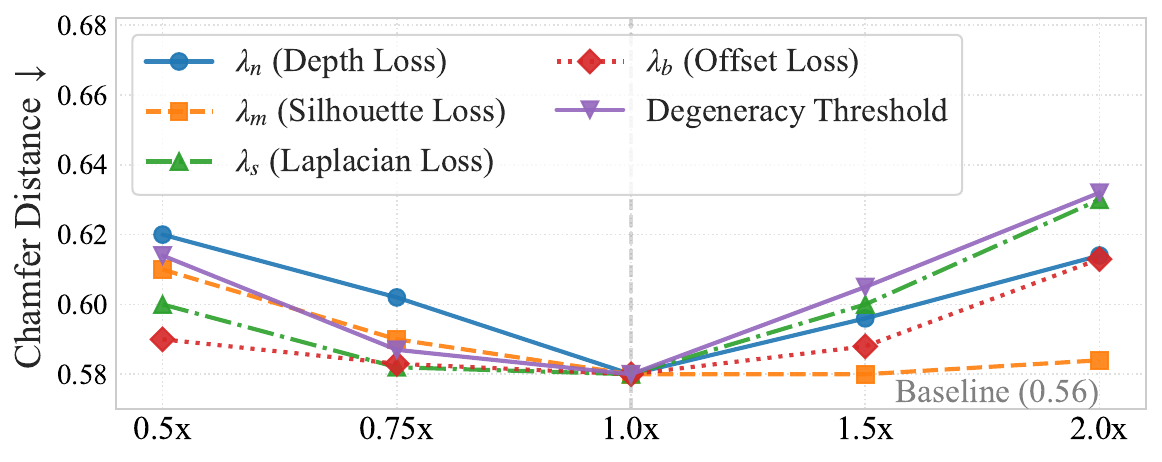}
    \vspace{-6mm}
    \caption{Ablation on hyperparameter sensitivity. The geometric accuracy remains highly stable as each parameter varies within $[0.5\times, 2.0\times]$ of its default value.}
    \label{fig:hyperparameter_sensitivity}
    \vspace{-4mm}
\end{figure}

\noindent\textbf{Hyperparameter Robustness.} Finally, we conduct a comprehensive ablation study on five key hyperparameters, including the degeneracy threshold and various loss weights. As shown in Figure~\ref{fig:hyperparameter_sensitivity}, the Chamfer Distance on the DTU dataset remains highly stable as each parameter varies within $[0.5\times, 2.0\times]$ of its default value.

\section{Conclusion}
\label{sec:conclusion}

In this paper, we propose ExMesh, an explicit mesh reconstruction framework designed to address the critical challenge of balancing high fidelity and conciseness. By integrating differentiable optimization with discrete topology updates and real-time UV maintenance, our method achieves coarse-to-fine mesh refinement. While our current framework demonstrates robustness across hyperparameter settings, exploring self-adaptive optimization mechanisms remains a promising direction for future work.
\section{Acknowledgments}
\label{sec:acknowledgment}

This work was supported by the Open Fund of National Key Laboratory of Deep Space Exploration (Grant NKDSEL2025008), the National Natural Science Foundation of China (Grant No. 62306294), and the Key Technology Research Project of TW-3 (TW3004).
{
    \small
    \FloatBarrier 
    \bibliographystyle{ieeenat_fullname}
    \bibliography{main}
}
\clearpage
\setcounter{page}{1}
\maketitlesupplementary

\newcommand{\M}{\mathcal{M}}
\newcommand{\V}{\mathcal{V}}
\newcommand{\F}{\mathcal{F}}
\newcommand{\U}{\mathcal{U}}
\newcommand{\Umap}{\mathcal{U}^{\text{map}}}
\newcommand{\Gv}{\mathcal{G}_{\mathcal{V}}}
\newcommand{\Kf}{\mathcal{K}}
\newcommand{\Gf}{\mathcal{G}}
\newcommand{\Sf}{S}
\newcommand{\Fc}{\mathcal{F}_{\text{cand}}}
\newcommand{\Fs}{\mathcal{F}_{\text{split}}}
\newcommand{\Cr}{\mathcal{C}_{\text{render}}}
\newcommand{\Df}{\mathcal{D}}
\newcommand{\Fm}{\mathcal{F}_{\text{merge}}}
\newcommand{\Fmod}{\mathcal{F}_{\text{mod}}}

\algrenewcommand\algorithmicrequire{\textbf{Input:}}
\algrenewcommand\algorithmicensure{\textbf{Output:}}
\algrenewcommand\algorithmiccomment[1]{\hfill\textit{// #1}}

\section{Methodology}
\label{sec:methodology}

\subsection{Pseudocode for Topology Operations}

Sections 3.2 and 3.3 introduce our core adaptive vertex splitting and merging strategies. This section provides detailed pseudocode for these operations and their corresponding UV maintenance routines.

\begin{algorithm}[H]
\caption{Adaptive Vertex Splitting}
\label{alg:split}
\begin{algorithmic}[1]
\Require 
    Mesh $\M(\V, \F)$, UV coordinates $\U$, UV map $\Umap$,
    Vertex EMA Gradients $\Gv$, Face Curvatures $\Kf$,
    Num splits $N_{\text{split}}$, Hyperparameters $\alpha, \beta, \tau_{\text{area\_hi}}, \tau_{\text{pos}}, \tau_{\text{uv}}$
\Ensure 
    Updated Mesh $\M'(\V', \F')$, Updated UVs $\U'$

\State $\Fc \gets \emptyset$ \Comment{Set of candidate faces}
\State $\Fmod \gets \emptyset$ \Comment{Set of modified faces}
\For{each face $\F_i \in \F$}
    \If{$Area(\F_i) \ge \tau_{\text{area\_hi}}$} 
        \State $\Gf_i \gets \frac{1}{3} \sum_{v \in \F_i} \Gv(v)$ 
        \State $\Sf_i \gets \alpha \Gf_i + \beta \Kf_i$ 
        \State $\Fc \gets \Fc \cup \{(\F_i, \Sf_i)\}$
    \EndIf
\EndFor

\State $\Fs \gets \text{SampleTopN}(\Fc, N_{\text{split}})$ 

\For{each face $\F(v_a, v_b, v_c) \in \Fs$}
    \If{$\F \in \Fmod$} \textbf{continue} \EndIf
    \State $e(v_a, v_b) \gets \text{LongestEdge}(\F)$
    \State $\F'(v_a, v_b, v_d) \gets \text{AdjacentFace}(\M, e)$ 
    
    \If{$\F' \neq \text{NULL}$} \Comment{Non-boundary edge}
        \If{$\F' \in \Fmod$} \textbf{continue} \EndIf 
        \State $t_c \gets \text{Project}(v_c, e)$; \quad $t_d \gets \text{Project}(v_d, e)$
        \State $v_s \gets (t_c + t_d) / 2$ 
        \State $\mu \gets \frac{\|v_s - v_a\|}{\|v_b - v_a\|}$ 
        \If{$\mu < \tau_{\text{pos}}$ \textbf{or} $\mu > 1 - \tau_{\text{pos}}$} \textbf{continue} \EndIf 
        
        \State $\V \gets \V \cup \{v_s\}$
        \State $\F \gets \F \setminus \{\F, \F'\}$ 
        \State $\F_{\text{new}} \gets \{\F(v_a, v_s, v_c), \F(v_s, v_b, v_c),$
        \State \hspace{3.5em} $\F(v_a, v_s, v_d), \F(v_s, v_b, v_d)\}$
        \State $\F \gets \F \cup \F_{\text{new}}$ 
        \State $\Fmod \gets \Fmod \cup \{\F, \F'\} \cup \F_{\text{new}}$
    \Else \Comment{Boundary edge}
        \State $v_s \gets (v_a + v_b) / 2$; \quad $\mu \gets 0.5$
        \State $\V \gets \V \cup \{v_s\}$
        \State $\F \gets \F \setminus \{\F\}$ 
        \State $\F_{\text{new}} \gets \{\F(v_a, v_s, v_c), \F(v_s, v_b, v_c)\}$
        \State $\F \gets \F \cup \F_{\text{new}}$ 
        \State $\Fmod \gets \Fmod \cup \{\F\} \cup \F_{\text{new}}$
    \EndIf
    \State $(\U, \Umap) \gets \textbf{HandleUVsForVertexSplit}(\cdot)$
\EndFor
\State $(\U, \Umap) \gets \textbf{CompactUVs}(\cdot)$ 
\State \Return $\M'(\V', \F'), \U', {\Umap}'$
\end{algorithmic}
\end{algorithm}

\begin{algorithm}[H]
\caption{Adaptive Vertex Merging}
\label{alg:merge}
\begin{algorithmic}[1]
\Require 
    Mesh $\M(\V, \F)$, UV coordinates $\U$, UV map $\Umap$,
    Face Render Counts $\Cr$, Hyperparameters $\tau_{\text{degen}}, \tau_{\text{area\_low}}$
\Ensure 
    Updated Mesh $\M'(\V', \F')$, Updated UVs $\U'$

\State $\Fm \gets \emptyset$ \Comment{Set of candidate faces}
\State $\Fmod \gets \emptyset$ \Comment{Set of modified faces}
\For{each face $\F_i \in \F$}
    \If{$Area(\F_i) \le \tau_{\text{area\_low}}$} 
        \State $\Df_i \gets Area(\F_i) / \text{LongestEdge}(\F_i)^2$ 
        \If{$\Cr(\F_i) = 0$ \textbf{or} $\Df_i < \tau_{\text{degen}}$} 
            \State $\Fm \gets \Fm \cup \{\F_i\}$
        \EndIf
    \EndIf
\EndFor

\For{each face $\F(v_i, v_j, v_m) \in \Fm$}
    \If{$\F \in \Fmod$} \textbf{continue} \EndIf
    
    \If{$\text{IsBoundaryFace}(\F)$}
        \State $e(v_i, v_m) \gets \text{GetBoundaryEdge}(\F)$ 
    \Else
        \State $e(v_i, v_m) \gets \text{ShortestEdge}(\F)$ 
    \EndIf
    
    \If{$Degree(v_m) > Degree(v_i)$} 
        \State $\text{swap}(v_i, v_m)$
    \EndIf
    
    \State $AdjFaces \gets \text{GetAdjacentFaces}(\M, v_m)$
    \If{$\text{CheckConflict}(AdjFaces, \Fmod)$} \textbf{continue} \EndIf 
    \For{each face $\F_k \in AdjFaces$}
        \If{$v_i \in \F_k$} \Comment{$\F_k$ contains both $v_i$ and $v_m$}
            \State $\F \gets \F \setminus \{\F_k\}$ 
        \Else \Comment{$\F_k$ only contains $v_m$}
            \State $\F_k \gets \text{ReplaceVertex}(\F_k, v_m, v_i)$ 
        \EndIf
        \State $\Fmod \gets \Fmod \cup \{\F_k\}$
    \EndFor
    \State $\V \gets \V \setminus \{v_m\}$ 
\EndFor

\If{$|\Fm| > 0$} 
    \State $(\U, \Umap) \gets \textbf{CompactUVs}(\cdot)$ 
\EndIf
\State \Return $\M'(\V', \F'), \U', {\Umap}'$
\end{algorithmic}
\end{algorithm}

\begin{algorithm}[H]
\caption{Compact UV}
\label{alg:uv_compact}
\begin{algorithmic}[1]
\Require 
    Mesh $\M(\V, \F)$, UVs $\U$, UV map $\Umap$
\Ensure 
    Compacted UVs $\U'$, Compacted UV map ${\Umap}'$

\State $S \gets$ set of all UV indices referenced in $\Umap$
\State $\U' \gets$ list of $\U[i]$ for $i \in S$, ordered by new index
\State Build $OldToNew$ mapping from old indices in $S$ to new indices in $\U'$
\For{each face $\F_i \in \F$}
    \State Update ${\Umap}'[\F_i]$ by replacing each old index with $OldToNew[\cdot]$
\EndFor
\State \Return $\U', {\Umap}'$
\end{algorithmic}
\end{algorithm}

\begin{algorithm}[H]
\caption{UV Interpolation for New Vertex}
\label{alg:uv_interpolation}
\begin{algorithmic}[1]
\Require 
    Vertices $\V$, Faces $\F$, UVs $\U$, UV map $\Umap$, New vertex $v_s$,
    Split edge $e(v_a, v_b)$, Old faces $\F_{\text{old}}, \F'_{\text{old}}$, Relative pos $\mu$, Seam threshold $\tau_{\text{uv}}$
\Ensure 
    Updated UVs $\U'$

\State $(u_a, u_b) \gets \text{GetUVsForFace}(\Umap, \F_{\text{old}}, e)$
\State $u_s^{(1)} \gets (1 - \mu)u_a + \mu u_b$ 

\If{$\F'_{\text{old}} = \text{NULL}$} \Comment{Boundary edge}
    \State $idx_s \gets \text{AppendToList}(\U, u_s^{(1)})$ 
    \State \textbf{for} each new face $\F_{\text{new}}$ containing $v_s$, assign $idx_s$ to $v_s$ in $\F_{\text{new}}$ and update $\Umap$
\Else \Comment{Non-boundary edge}
    \State $(u'_a, u'_b) \gets \text{GetUVsForFace}(\Umap, \F'_{\text{old}}, e)$
    \State $u_s^{(2)} \gets (1 - \mu)u'_a + \mu u'_b$ 
    
    \If{$\|u_s^{(1)} - u_s^{(2)}\| < \tau_{\text{uv}}$} \Comment{Check for UV seam}
        \State $u_s \gets (u_s^{(1)} + u_s^{(2)}) / 2$ \Comment{no seam}
        \State $idx_s \gets \text{AppendToList}(\U, u_s)$
        \State \textbf{for} each new face $\F_{\text{new}}$ containing $v_s$, assign $idx_s$ to $v_s$ in $\F_{\text{new}}$ and update $\Umap$
    \Else \Comment{Create a new UV seam}
        \State $idx_s^{(1)} \gets \text{AppendToList}(\U, u_s^{(1)})$
        \State $idx_s^{(2)} \gets \text{AppendToList}(\U, u_s^{(2)})$
        \State \textbf{for} each new face $\F_{\text{new}}$ containing $v_s$, assign $idx_s^{(1)}$ or $idx_s^{(2)}$ to $v_s$, and update $\Umap$
    \EndIf
\EndIf
\State \Return $\U', {\Umap}'$
\end{algorithmic}
\end{algorithm}

\subsection{Loss Function}
As described in Section 3.5, the overall optimization objective is formulated as a weighted sum of several loss terms. Below, we detail the main components of the loss function used in our framework.

\noindent\textbf{Photometric Loss} Following the formulation in 3D Gaussian Splatting~\cite{kerbl2023gaussian}, our primary photometric loss is a combination of an $\mathcal{L}_{1}$ loss and a D-SSIM (Structural Dissimilarity) term, which balances absolute color accuracy with perceptual similarity:
\begin{equation}
\mathcal{L}_{\text{rgb}} = (1 - \lambda_{\text{dssim}})\mathcal{L}_{1}(\hat{I}, I_{\text{gt}}) + \lambda_{\text{dssim}}\mathcal{L}_{\text{D-SSIM}}(\hat{I}, I_{\text{gt}})
\end{equation}
Here, $\hat{I}$ is the rendered RGB image and $I_{\text{gt}}$ is the GT image. The $\mathcal{L}_{1}$ term provides a robust pixel-wise difference, while the D-SSIM term~\cite{wang2004ssim} adds a perceptually-driven gradient.

\noindent\textbf{Depth Consistency Loss} To ensure that the rendered surface geometry is consistent with the reference depth, we adopt a Pearson depth loss. This loss penalizes the discrepancy between the rendered depth map $\hat{D}$ and the reference depth $D_{\text{ref}}$ obtained from Depth Anything 3~\cite{lin2025depthanything3}:
\begin{equation}
\mathcal{L}_{d} = 1 - \frac{\text{Cov}(\hat{D}, D_{\text{ref}})}{\sigma_{\hat{D}} \sigma_{D_{\text{ref}}}}
\end{equation}
where $\text{Cov}$ denotes the covariance and $\sigma$ denotes the standard deviation, both computed within the silhouette mask.

\begin{table}[t]
    \centering
    \caption{Key hyperparameters used in experiments.}
    \label{tab:hyperparams}
    \resizebox{\columnwidth}{!}{
    \begin{tabular}{lcc}
        \toprule
        \textbf{Parameter} & \textbf{Symbol} & \textbf{Value} \\
        \midrule
        Vertex learning rate & $lr_{\text{vertices}}$ & 0.0005 \\
        Feature learning rate & $lr_{\text{feature}}$ & 0.0025 \\
        Photometric loss weight & $\lambda_{\text{rgb}}$ & 1.0 \\
        D-SSIM loss weight & $\lambda_{\text{dssim}}$ & 0.2 \\
        Depth loss weight & $\lambda_{d}$ & 0.01 \\
        Silhouette loss weight & $\lambda_{m}$ & 0.005 \\
        Laplacian loss weight & $\lambda_{s}$ & 1000 \\
        Offset loss weight & $\lambda_{b}$ & 1000 \\
        Area threshold (split) & $\tau_{\text{area\_hi}}$ & 1.5$\times$ median \\
        Area threshold (merge) & $\tau_{\text{area\_low}}$ & 0.5$\times$ median \\
        Degeneracy threshold & $\tau_{\text{degen}}$ & 0.05 \\
        Texture resolution & $-$ & $2560 \times 2560$ \\
        \bottomrule
    \end{tabular}
    }
\vspace{-2mm}
\end{table}

\noindent\textbf{Silhouette Loss} Additionally, a standard silhouette loss is employed to constrain the mesh geometry to the object's boundaries. This loss is formulated as a Binary Cross-Entropy (BCE) between the rendered alpha mask $\hat{M}$ and the ground-truth 2D mask $M_{\text{gt}}$:
\begin{equation}
\mathcal{L}_{m} = \text{BCE}(\hat{M}, M_{\text{gt}})
\end{equation}

\noindent\textbf{Laplacian Smoothing Loss} 
A Laplacian smoothing loss~\cite{desbrun1999implicit,taubin1995signal} is further incorporated to promote a smooth and well-formed mesh topology. By minimizing the discrete Laplacian vector at each vertex $v_i$ relative to its 1-ring neighbors $\mathcal{N}(i)$, this regularizer penalizes high-frequency noise and encourages uniform vertex distribution:
\begin{equation}
\mathcal{L}_{s} = \frac{1}{|\mathcal{V}|} \sum_{v_i \in \mathcal{V}} \left\| \left( v_i - \frac{1}{|\mathcal{N}(i)|} \sum_{v_j \in \mathcal{N}(i)} v_j \right) \right\|^2
\end{equation}

\noindent\textbf{Bi-vertex Offset Loss} Similar to FlexiCubes~\cite{shen2023flexicubes}, ExMesh operates on a deformable grid where each vertex $v_i$ has a deformation offset $\Delta v_i$. To prevent overly sharp or disjointed deformations between adjacent grid vertices, a bi-vertex offset regularization loss is introduced, which encourages the deformation field to remain locally smooth.
\begin{equation}
\mathcal{L}_{b} = \frac{1}{|\mathcal{V}|} \sum_{v_i \in \mathcal{V}} \left\| \left( \Delta v_i - \frac{1}{|\mathcal{N}(i)|} \sum_{v_j \in \mathcal{N}(i)} \Delta v_j \right) \right\|^2
\end{equation}

\begin{figure*}[t]
    \centering
    \includegraphics[width=\linewidth]{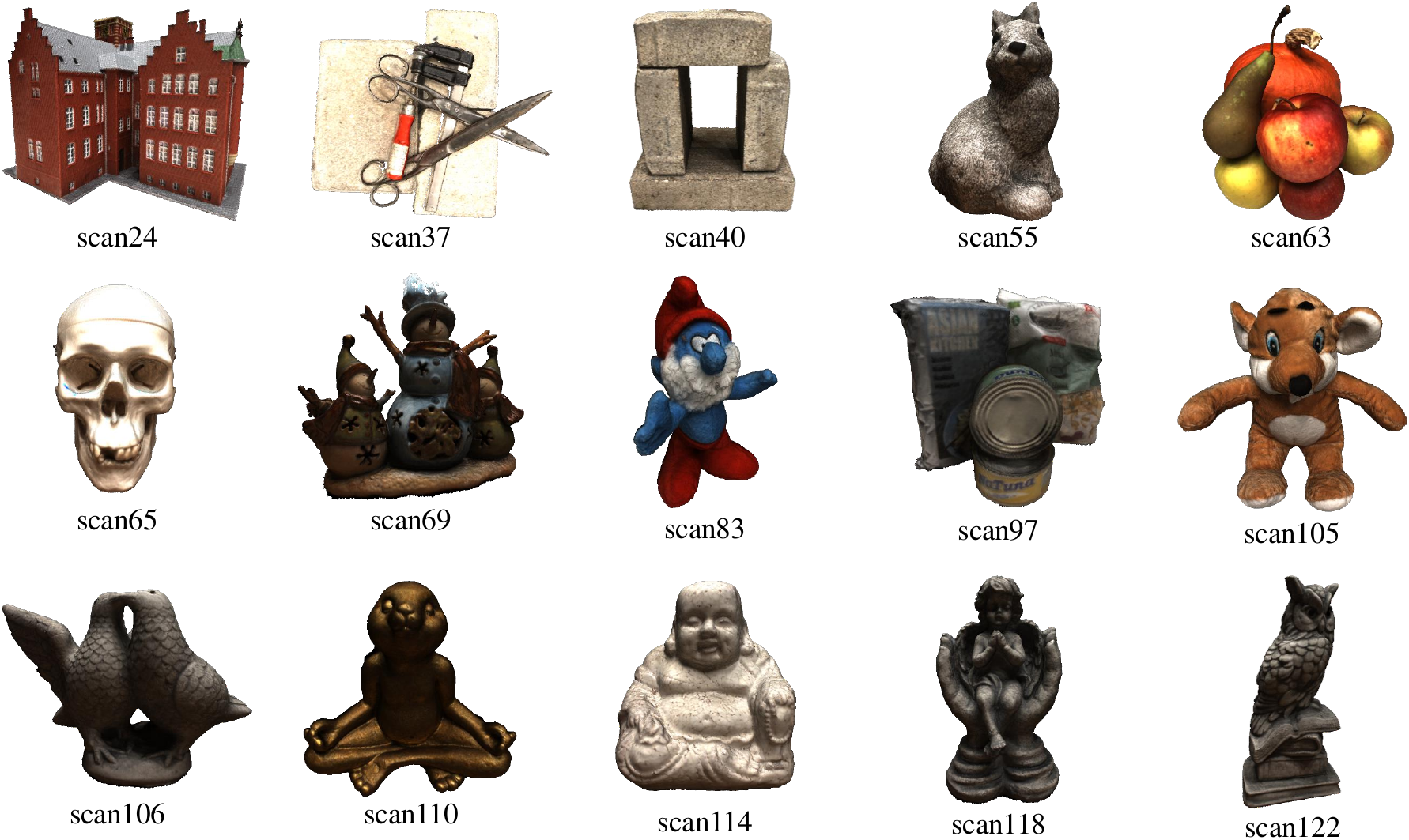}
    \caption{Visualization of Reconstructed Meshes with Vertex Color on DTU Dataset.}
    \label{fig:sup_colored_DTU}
\end{figure*}

\section{Implementation Details}
\subsection{Hyperparameters}

Table~\ref{tab:hyperparams} summarizes the main hyperparameters used in our framework. All values are selected based on validation performance and follow common practice in related works.

\subsection{Baselines}

All baselines are trained using their official implementations and recommended settings. For methods that do not produce an explicit mesh, surfaces are extracted using the authors' prescribed procedures (e.g., TSDF fusion or Marching Cubes) to ensure a fair comparison. Note that consistent with our method, we apply object masks during the evaluation of all baselines to filter out background noise and focus on the reconstruction quality of the target object.

\noindent\textbf{NeRF-based (Implicit) Methods.}
\textbf{VolSDF}~\cite{yariv2021volsdf} represents the surface as a neural SDF and defines the volume density using Laplace's cumulative distribution function.
\textbf{NeuS}~\cite{wang2021neus} also represents the surface as a neural SDF, but introduces an unbiased S-density function for volume rendering.
\textbf{Neuralangelo}~\cite{tancik2023neuralangelo} scales implicit surface reconstruction to large scenes by encoding the SDF in multi-resolution hash grids and employs a coarse-to-fine optimization schedule with geometric regularization.

\noindent\textbf{GS-based (Explicit) Methods.}
\textbf{SuGaR}~\cite{su2023sugar} introduces a regularization term to encourage 3D Gaussians to align with surfaces and extracts a coarse mesh via Poisson reconstruction.
\textbf{2DGS}~\cite{huang2024twodgs} represents the scene using explicit 2D Gaussian disks, providing stronger geometric constraints. The final mesh is extracted using TSDF fusion.
\textbf{PGSR}~\cite{chen2024pgsr} uses planar primitives derived from 3D Gaussians, introduces an unbiased depth rendering method, and enforces geometric consistency through single-view and multi-view regularization.
\textbf{Triangle Splatting}~\cite{held2025trianglesplatting} proposes a differentiable renderer that directly optimizes an unstructured triangle soup, enabling end-to-end optimization of vertex positions and colors.
\textbf{GOF}~\cite{yu2024gof} derives a continuous opacity field from optimized Gaussians, enabling surface extraction by identifying a level-set and using Marching Tetrahedra.
\textbf{QGS}~\cite{zhang2025qgs} extends 3DGS by introducing second-order quadric primitives, allowing each primitive to represent more complex local geometry.

\noindent\textbf{Mesh-driven (Hybrid) Methods.}
\textbf{Nvdiffrec}~\cite{munkberg2022nvdiffrec} jointly optimizes topology, materials, and lighting from images. It uses a differentiable marching tetrahedra (DMTet) layer to extract a mesh from an intermediate volumetric SDF grid inside the training loop.
\textbf{FlexiCubes}~\cite{shen2023flexicubes} also optimizes an SDF grid, and introduces a differentiable isosurface extraction layer based on Dual Marching Cubes, with learnable parameters for vertex positioning and quad splitting.
\textbf{IMLS-Splatting}~\cite{yang2025imlssplatting} uses a point cloud as the core representation and introduces a differentiable Implicit Moving-Least Squares (IMLS) algorithm to generate a sparse SDF grid for meshing.
\textbf{GeoSVR}~\cite{li2025geosvr} is based on a sparse voxel representation, which is optimized directly with monocular depth constraints and surface regularization.

\vspace{-2mm}
\section{Additional Results}
\subsection{Additional Results on DTU Dataset}
\begin{figure*}[t]
    \centering
    \includegraphics[width=0.92\linewidth]{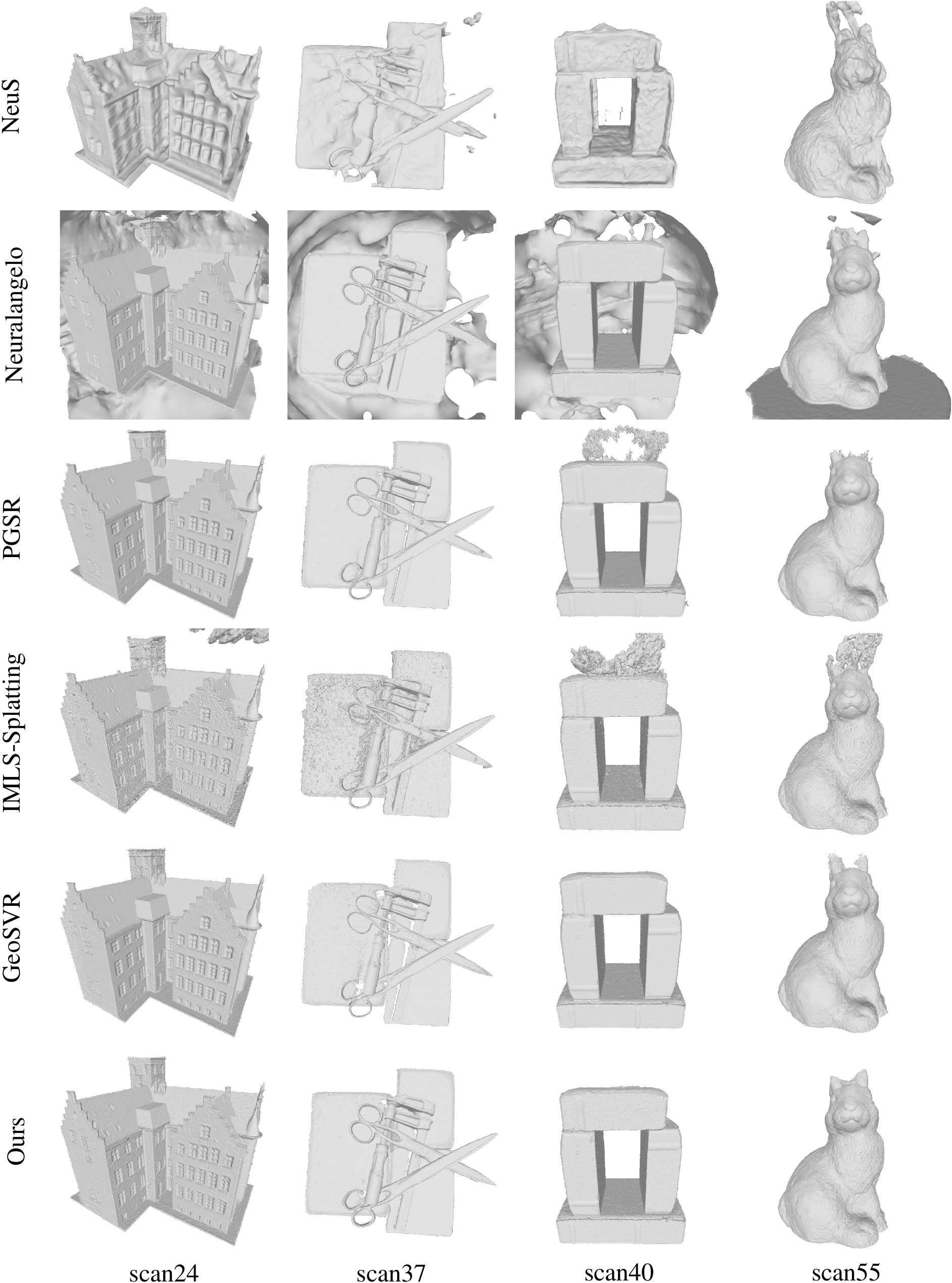}
    \vspace{2mm}
    \caption{Qualitative geometric comparison on DTU dataset (part 1).}
    \label{fig:sup_DTU1}
\end{figure*}

\begin{figure*}[t]
    \centering
    \includegraphics[width=0.92\linewidth]{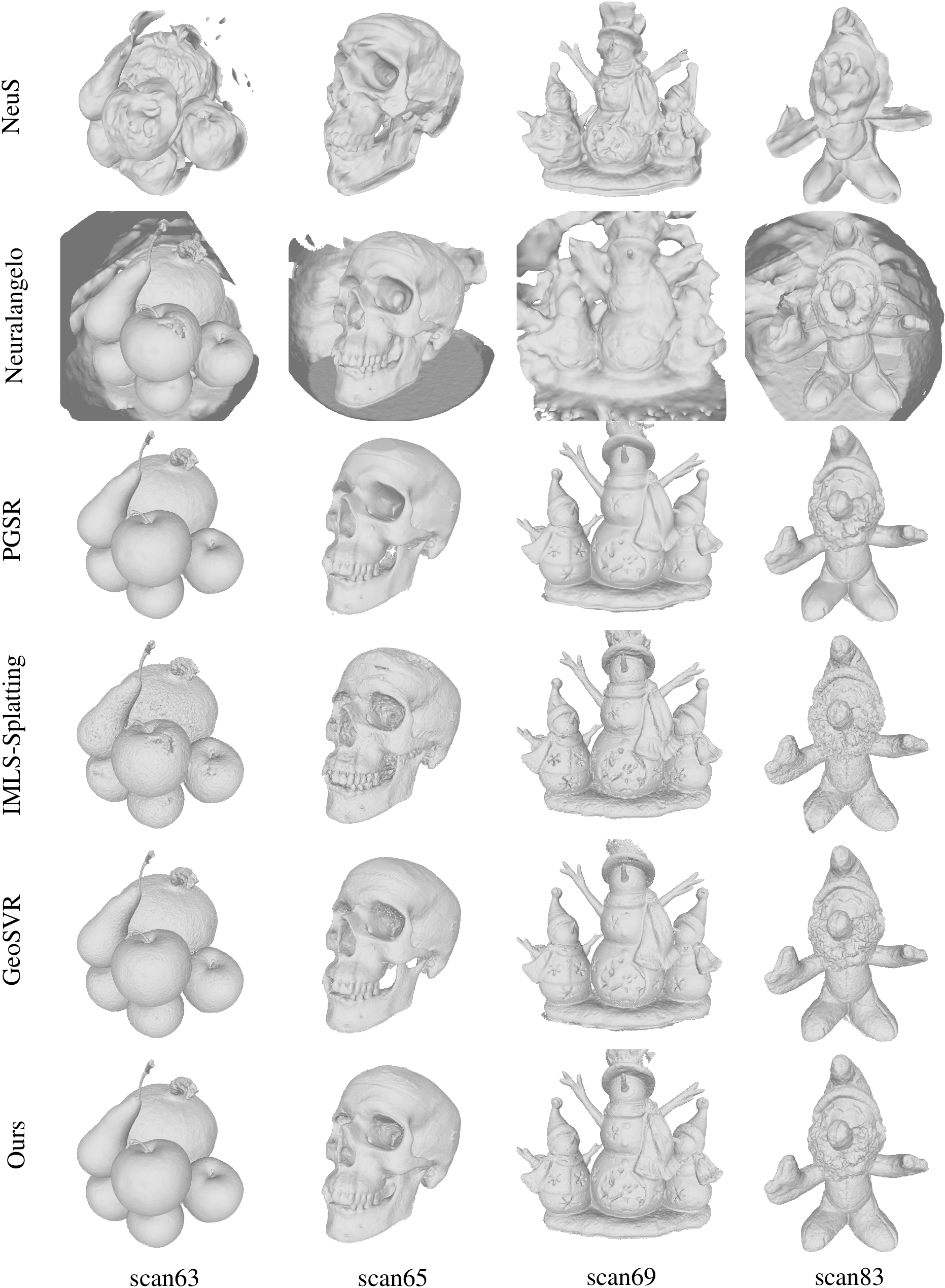}
    \vspace{2mm}
    \caption{Qualitative geometric comparison on DTU dataset (part 2).}
    \label{fig:sup_DTU2}
\end{figure*}

\begin{figure*}[t]
    \centering
    \includegraphics[width=0.92\linewidth]{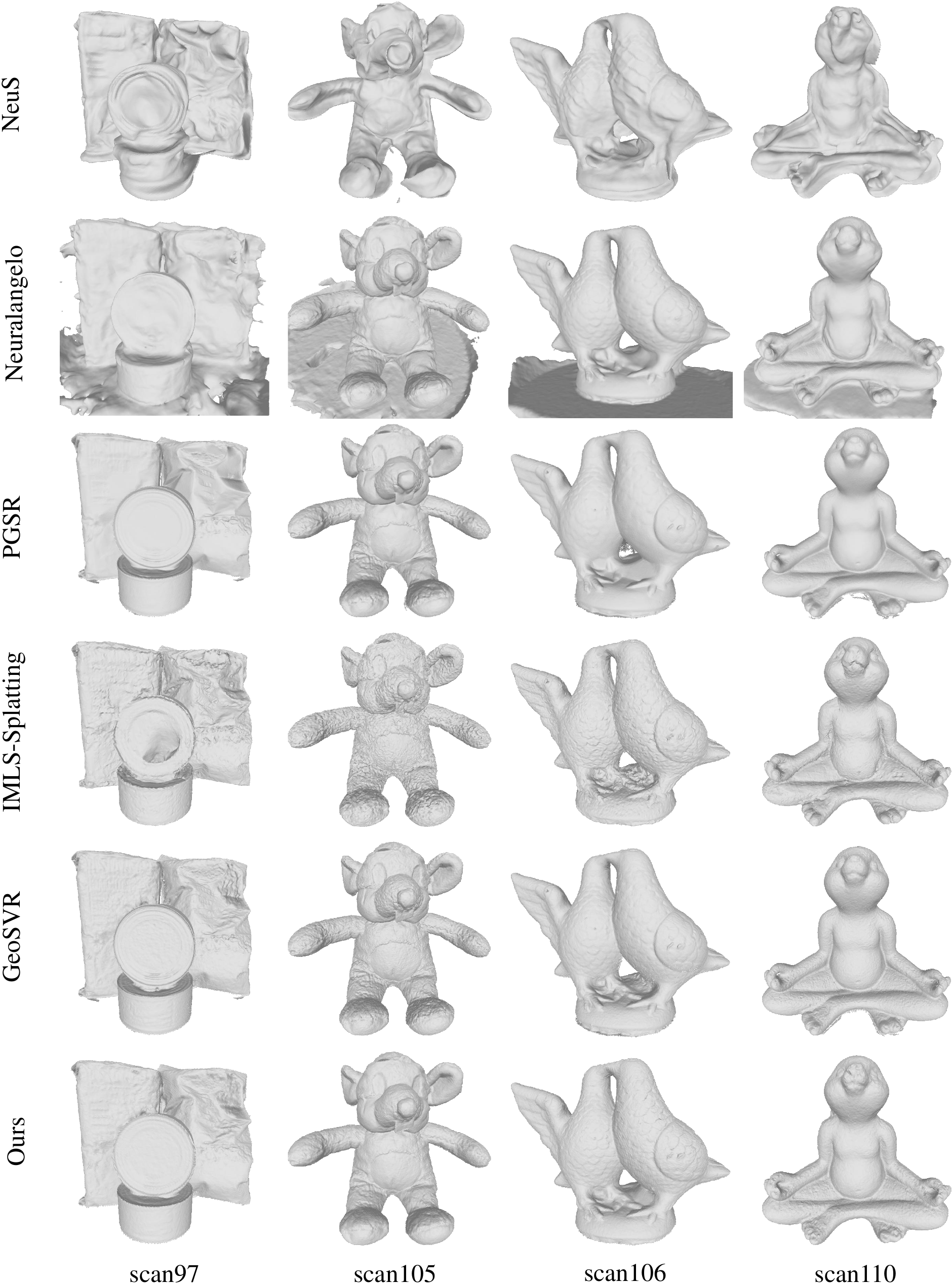}
    \vspace{2mm}
    \caption{Qualitative geometric comparison on DTU dataset (part 3).}
    \label{fig:sup_DTU3}
\end{figure*}

\begin{figure*}[t]
    \centering
    \includegraphics[width=0.7\linewidth]{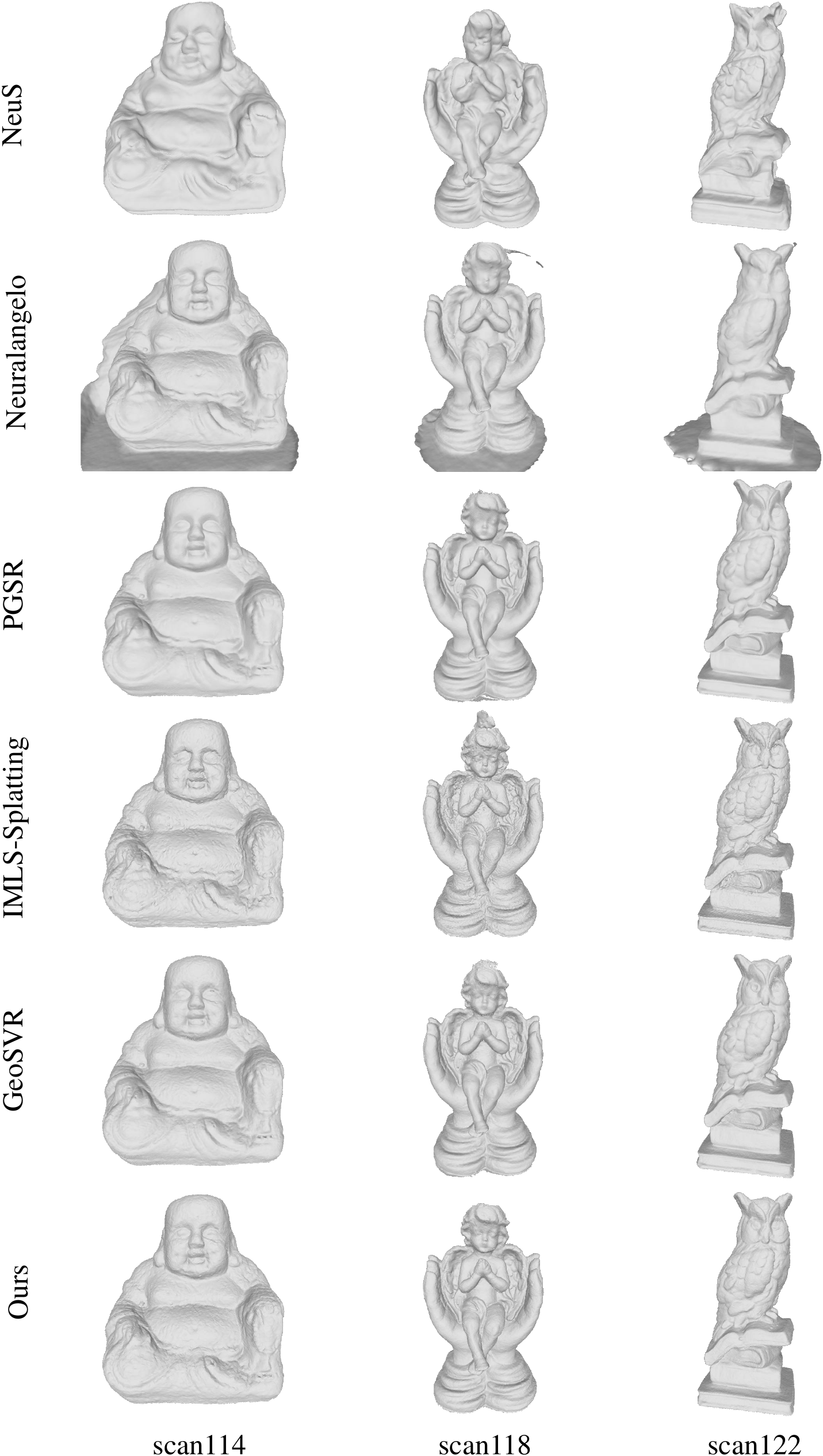}
    \vspace{2mm}
    \caption{Qualitative geometric comparison on DTU dataset (part 4).}
    \label{fig:sup_DTU4}
\end{figure*}

Figure~\ref{fig:sup_colored_DTU} shows meshes with vertex color reconstructed by our method on the DTU \cite{jensen2014dtu} dataset, demonstrating high-fidelity geometry and realistic surface appearance. In addition, Figures~\ref{fig:sup_DTU1}--\ref{fig:sup_DTU4} present a comprehensive visual comparison between our approach and several recent baselines, including NeuS~\cite{wang2021neus}, Neuralangelo~\cite{tancik2023neuralangelo}, PGSR~\cite{chen2024pgsr}, IMLS-Splatting~\cite{yang2025imlssplatting}, and GeoSVR~\cite{li2025geosvr}. ExMesh achieves more accurate and complete surface reconstruction, with fewer artifacts and better preservation of fine details, highlighting its robustness and effectiveness on real-world objects.

\subsection{Efficiency Analysis}

We compare the efficiency of our method with recent baselines in terms of total reconstruction time and peak GPU memory usage on the DTU dataset \cite{jensen2014dtu}. As summarized in Table~\ref{tab:efficiency}, our method achieves competitive or superior efficiency, requiring significantly less training time and memory compared to most approaches. These results demonstrate the practical advantages of our framework for scalable and resource-efficient surface reconstruction.

\begin{table}[t]
    \centering
    \caption{Efficiency comparison on DTU dataset. We report total reconstruction time and peak GPU memory usage.}
    \vspace{-1mm}
    \label{tab:efficiency}
    \resizebox{0.85\columnwidth}{!}{
    \begin{tabular}{lcc}
        \toprule
        \textbf{Method} & \textbf{Time $\downarrow$} & \textbf{Memory $\downarrow$} \\
        \midrule
        VolSDF~\cite{yariv2021volsdf}           & $>$12h   & 14GB   \\
        NeuS~\cite{wang2021neus}                & $>$12h   & 7GB   \\
        Neuralangelo~\cite{tancik2023neuralangelo} & $>$12h   &  16GB  \\
        SuGaR~\cite{su2023sugar}                & 1h   & 23GB   \\
        2DGS~\cite{huang2024twodgs}             & \cellcolor{colorfirst}{11m}   & 5GB   \\
        TriSplat~\cite{held2025trianglesplatting} & \cellcolor{colorsecond}{15m}   & 23GB   \\
        GOF~\cite{yu2024gof}                    & 1h   & 9GB   \\
        PGSR~\cite{chen2024pgsr}                & 30m   & \cellcolor{colorfirst}{3.5GB}   \\
        QGS~\cite{zhang2025qgs}                 & 48m   & \cellcolor{colorfirst}{3.5GB}   \\
        Nvdiffrec~\cite{munkberg2022nvdiffrec}  & $>$1h   &  20GB  \\
        IMLS-Splatting~\cite{yang2025imlssplatting} & \cellcolor{colorsecond}{15m}   &  20GB  \\
        GeoSVR~\cite{li2025geosvr}              & 49m   & 14GB   \\
        Ours                           & \cellcolor{colorsecond}{15m}   & \cellcolor{colorthird}{4.5GB}   \\
        \bottomrule
    \end{tabular}
    }
\vspace{-3mm}
\end{table}

\section{Additional Ablations}
\subsection{Loss Function}
\begin{figure}[t]
    \centering
    \includegraphics[width=\linewidth]{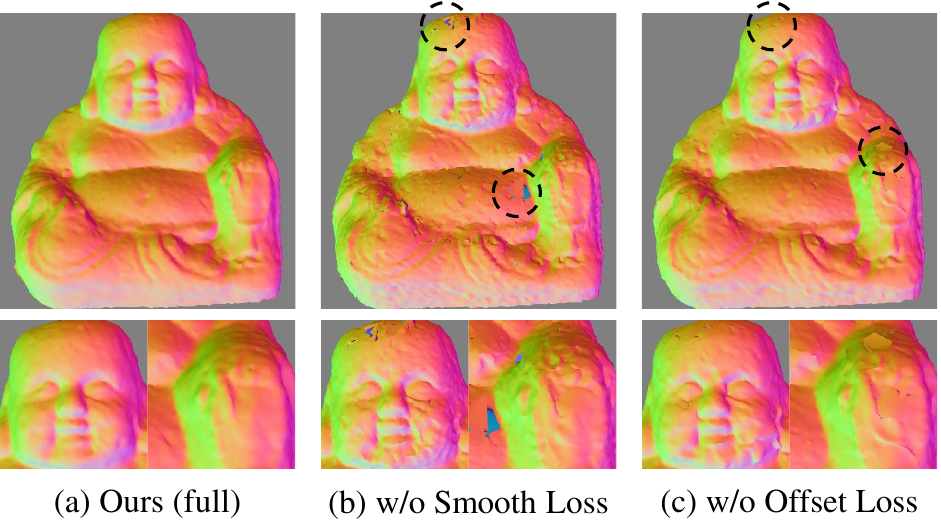}
    \caption{Ablation study on the Laplacian smoothing loss ($\mathcal{L}_{s}$) and bi-vertex offset loss ($\mathcal{L}_{b}$) on DTU dataset.}
    \label{fig:sup_ablation_loss}
\vspace{-3mm}
\end{figure}

To further investigate the impact of regularization terms in our framework, we conduct ablation studies on the Laplacian smoothing loss ($\mathcal{L}_{s}$) and the bi-vertex offset regularization loss ($\mathcal{L}_{b}$) on the DTU dataset. As illustrated in Figure~\ref{fig:sup_ablation_loss}, removing $\mathcal{L}_{s}$ leads to increased high-frequency noise and jagged artifacts, especially along thin structures and boundaries, indicating its importance for suppressing local fluctuations and maintaining mesh uniformity. In contrast, omitting $\mathcal{L}_{b}$ results in discontinuities and abrupt changes in the deformation field, causing locally stretched or folded regions and compromising surface regularity.

\begin{figure}[t]
    \centering
    \includegraphics[width=0.95\linewidth]{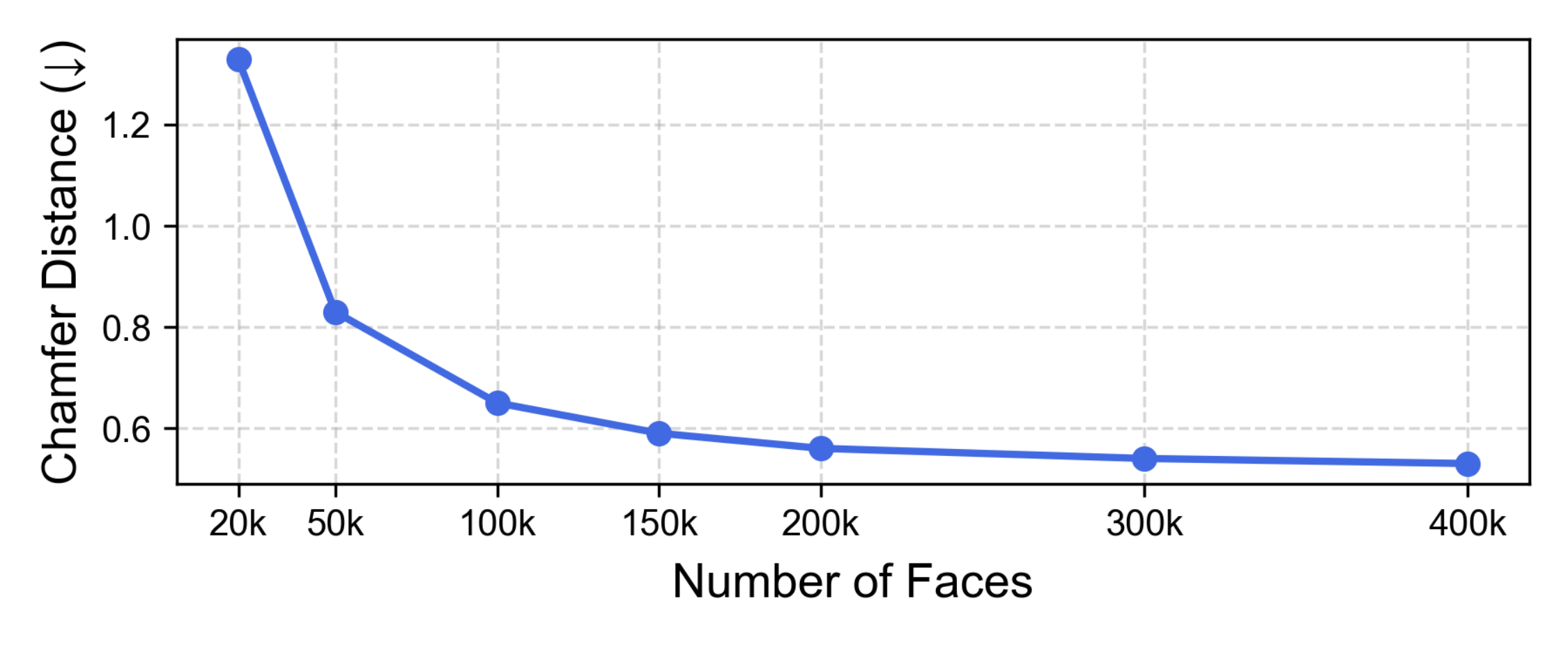}
    \vspace{-4mm}
    \caption{Analysis of the relationship between geometric accuracy and the final mesh face count on the DTU dataset.}
    \label{fig:ablation_faces}
    \vspace{-4mm}
\end{figure}

\subsection{Effect of Face Count on Performance}
Finally, we analyze the impact of face count on geometric accuracy. As shown in Figure~\ref{fig:ablation_faces}, geometric accuracy improves significantly as face count increases to 100k, but saturates beyond 200k. This indicates that further increasing face count yields little benefit once the mesh is sufficiently complex. Moreover, excessive faces significantly rendering time, making 150k-250k an optimal balance between quality and efficiency.

\section{Discussion}
While our method achieves strong reconstruction quality and efficiency, there remain practical limitations. First, the framework depends on careful tuning of multiple hyperparameters, such as loss weights and topology update thresholds, which could be time-consuming and may require expert knowledge. In addition, the efficiency of UV unwrapping poses a bottleneck for large meshes, as current tools like xatlas are CPU-based and lack CUDA acceleration. When the face count exceeds 200,000, UV unwrapping becomes significantly slower, limiting the scalability of our approach for scene-level reconstruction. Addressing these challenges, including developing automated parameter selection strategies and exploring GPU-accelerated UV unwrapping, will be an important direction for future work.

\end{document}